\documentclass[final]{cvpr}
\let\depth\relax

\usepackage{amssymb}
\usepackage{times}
\usepackage{epsfig}
\usepackage{graphicx}
\usepackage{amsmath}
\usepackage{mathabx}

\usepackage{booktabs}
\usepackage{multirow}
\usepackage[dvipsnames]{xcolor}
\usepackage{pifont}
\usepackage{wrapfig}
\usepackage[sort,nocompress]{cite}

\newcommand{\methodname}{PCME\xspace}
\newcommand{\pcmefull}{Probabilistic Cross-Modal Embedding\xspace}
\newcommand{\pcme}{\methodname}

\definecolor{darkergreen}{RGB}{21, 152, 56}
\definecolor{red2}{RGB}{252, 54, 65}

\newcommand{\cmark}{{\color{darkergreen} \ding{51}}}%
\newcommand{\xmark}{{\color{red2} \ding{55}}}%

\newcommand{\cD}{\mathcal{D}}
\newcommand{\cG}{\mathcal{G}}
\newcommand{\cI}{\mathcal{I}}
\newcommand{\cC}{\mathcal{C}}
\newcommand{\cV}{\mathcal{V}}
\newcommand{\cT}{\mathcal{T}}

\newcommand{\visencoder}{f_\cV}
\newcommand{\textencoder}{f_\cT}

\newcommand{\visfeat}{g_\cV}
\newcommand{\vishead}{h_\cV}
\newcommand{\textfeat}{g_\cT}
\newcommand{\texthead}{h_\cT}
\newcommand{\generichead}{h}

\usepackage[inline, shortlabels]{enumitem}
\setlist[itemize]{leftmargin=*, noitemsep, topsep=0pt}
\setlist[enumerate,1]{leftmargin=*, noitemsep, topsep=0pt, label={\bf (\roman*)}}

\newcommand{\vf}{v}
\newcommand{\tf}{t}

\newcommand{\vzf}{z_v}
\newcommand{\tzf}{z_t}

\newcommand{\att}{\texttt{att}}
\newcommand{\layernorm}{\texttt{LN}}

\newcommand{\realnum}{\mathbb{R}}
\newcommand{\sigmoid}{s}

\newcommand{\fcweight}{w}

\newcommand{\fref}[1]{\textcolor{red}{#1}}

\usepackage[pagebackref=true,breaklinks=true,colorlinks,bookmarks=false]{hyperref}

\def\cvprPaperID{274} %
\def\confYear{CVPR 2021}

\begin{document}

\title{Probabilistic Embeddings for Cross-Modal Retrieval}

\author{
Sanghyuk Chun$^1$~~Seong Joon Oh$^1$~~Rafael Sampaio de Rezende$^2$~~Yannis Kalantidis$^2$~~Diane Larlus$^2$
\\
\\
$^1$NAVER AI Lab \qquad $^2$NAVER LABS Europe
}

\maketitle

\begin{abstract}
    Cross-modal retrieval methods build a common representation space for samples from multiple modalities, typically from the vision and the language domains. For images and their captions, the multiplicity of the correspondences makes the task particularly challenging. Given an image (respectively a caption), there are multiple captions (respectively images) that equally make sense. In this paper, we argue that deterministic functions are not sufficiently powerful to capture such one-to-many correspondences. Instead, we propose to use \pcmefull (\pcme), where samples from the different modalities are represented as probabilistic distributions in the common embedding space. Since common benchmarks such as COCO suffer from non-exhaustive annotations for cross-modal matches, we propose to additionally evaluate retrieval on the CUB dataset, a smaller yet clean database where all possible image-caption pairs are annotated. We extensively ablate \pcme and demonstrate that it not only improves the retrieval performance over its deterministic counterpart but also provides uncertainty estimates that render the embeddings more interpretable. Code is available at \url{https://github.com/naver-ai/pcme}.
\end{abstract}

\section{Introduction}
\label{introduction}

\begin{figure}
    \centering
    \includegraphics[width=\columnwidth]{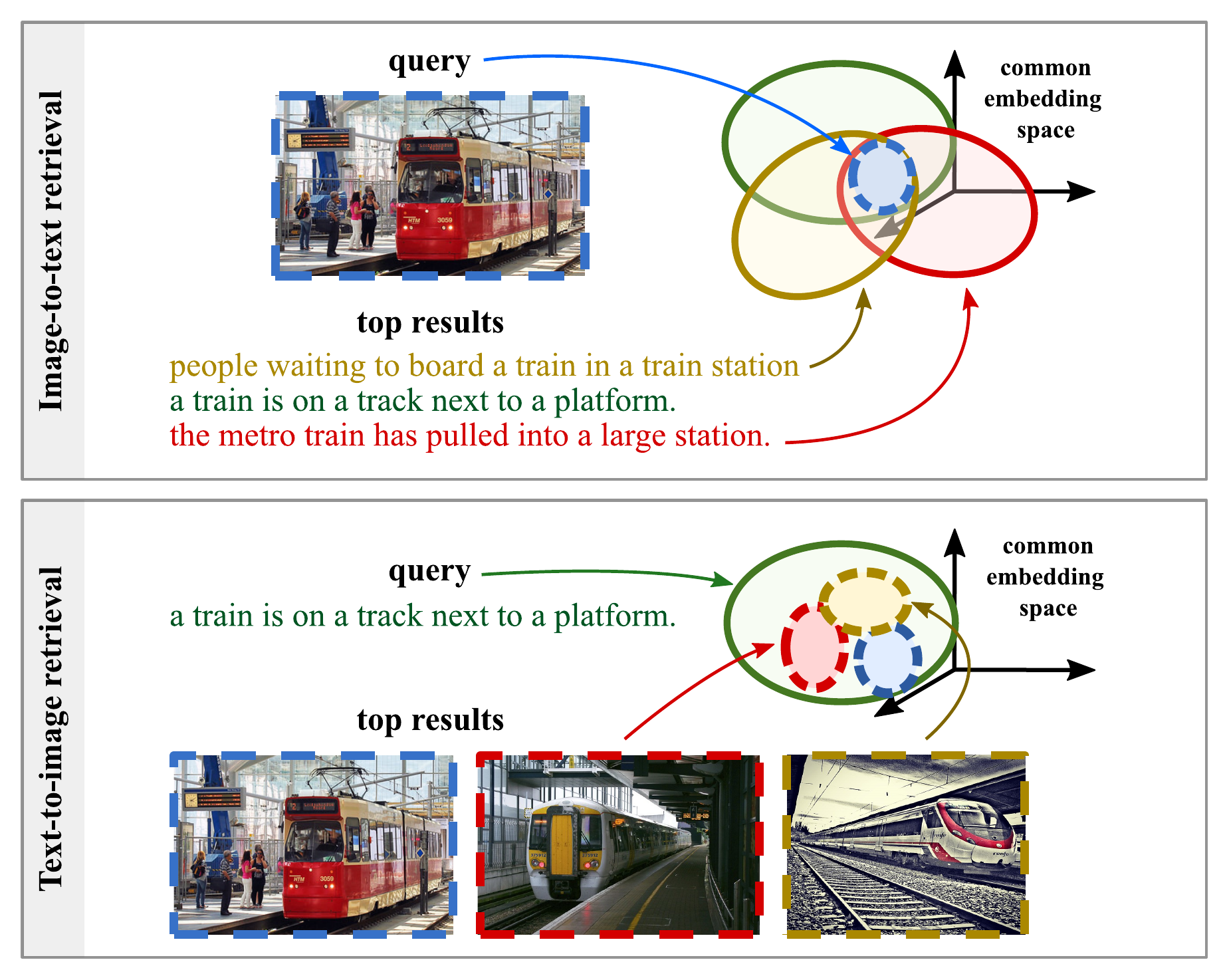} \\
    \caption{We propose to use \textbf{probabilistic embeddings} to represent images and their captions as probability distributions in a common embedding space suited for cross-modal retrieval. These distributions gracefully model the uncertainty which results from the multiplicity of concepts appearing in a visual scene and implicitly perform  many-to-many matching between those concepts.}
    \label{fig:teaser}
    \vspace{-.5em}
\end{figure}

Given a query and a database from different modalities, cross-modal retrieval is the task of retrieving the database items which are most relevant to the query. Most research on this topic has focused on the image and text modalities~\cite{chen2015microsoft,faghri2018vsepp,Li2019VSRN,wang2016learning,young2014image}. Typically, methods estimate embedding functions that map visual and textual inputs into a common embedding space, such that the cross-modal retrieval task boils down to the familiar nearest neighbour retrieval task in a Euclidean space~\cite{faghri2018vsepp,wang2016learning}. 

Building a common representation space for multiple modalities is challenging. Consider an image with a group of people on a platform preparing to board a train (Figure~\ref{fig:teaser}). There is more than one possible caption describing this image. ``People waiting to board a train in a train platform'' and ``The metro train has pulled into a large station'' were two of the choices from the COCO~\cite{chen2015microsoft} annotators. Thus, the common representation has to deal with the fact that an image potentially matches with a number of different captions. Conversely, given a caption, there may be multiple manifestations of the caption in visual forms. The multiplicity of correspondences across image-text pairs stems in part from the different natures of the modalities. All the different components of a visual scene are thoroughly and passively captured in a photograph, while language descriptions are the product of conscious choices of the key relevant concepts to report from a scene. All in all, a common representation space for image and text modalities is required to model the one-to-many mappings in both directions.

Standard approaches which rely on vanilla functions do not meet this necessary condition: they can only quantify one-to-one relationships~\cite{faghri2018vsepp,wang2016learning}. There have been attempts to introduce multiplicity. For example, Song and Soleymani~\cite{song2019pvse} have introduced Polysemous Visual-Semantic Embeddings (PVSE) by letting an embedding function propose $K$ candidate representations for a given input. PVSE has been shown to successfully capture the multiplicity in the matching task and to improve over the baseline built upon one-to-one functions. Others~\cite{Li2019VSRN} have computed region embeddings obtained with a pre-trained object detector, %
establishing multiple region-word matches. This strategy has led to significant performance gains at the expense of a significant increase in computational cost.

In this work, we propose \textbf{\pcmefull (\pcme)}. We argue that probabilistic mapping is an effective representation tool that does not require an explicit many-to-many representation as is done by detection-based approaches, and further offers a number of advantages. First, \pcme yields uncertainty estimates that lead to useful applications like estimating the difficulty or chance of failure for a query. Second, the probabilistic representation leads to a richer embedding space where set algebras make sense, whereas deterministic ones can only represent similarity relations. Third, \pcme is complementary to the deterministic retrieval systems.

As harmful as the assumption of one-to-one correspondence is for the method, the same assumption has introduced confusion in the evaluation benchmarks. For example, MS-COCO~\cite{chen2015microsoft} suffers from non-exhaustive annotations for cross-modal matches. The best solution would be to explicitly and manually annotate all image-caption pairs for evaluation. Unfortunately, this process does not scale, especially for a large-scale dataset like COCO. Instead, we propose a smaller yet cleaner cross-modal retrieval benchmark using CUB~\cite{welinder2010caltech} and more sensible evaluation metrics. 

Our contributions are as follows. (1) We propose \pcmefull (\pcme) to properly represent the one-to-many relationships in joint embedding spaces for cross-modal retrieval. (2) We identify shortcomings with existing cross-modal retrieval benchmarks and propose alternative solutions. (3) We analyse the joint embedding space using the uncertainty estimates provided by \pcme and show how intuitive properties arise. 

\section{Related work}
\label{sec:relatedwork}

\noindent\textbf{Cross-modal retrieval.}
In this work, we are interested in 
image and text
cross-modal retrieval. Much research is dedicated to learning a metric space that jointly embeds images and sentences~\cite{BeansBurgers,faghri2018vsepp,frome2013devise,karpathy2014deep,Li2019VSRN,song2019pvse,thomas2020preserving}. Early works~\cite{gong2014improving,klein2014fisher} relied on Canonical Correlation Analysis (CCA)~\cite{hardoon2004canonical} to build joint embedding spaces.
Frome \etal~\cite{frome2013devise} use a hinge rank loss for triplets built from both modalities. Wang \etal \cite{wang2016learning} expand on this idea by also training on uni-modal triplets to preserve the structure inherent to each modality in the joint space. Faghri \etal~\cite{faghri2018vsepp} propose to learn such space with a triplet loss, and only sample the hardest negative with respect to a query-positive pair.

One of the drawbacks of relying on a single global representation is its inability to represent the diversity of semantic concepts present in an image or in a caption. Prior work~\cite{huang2017instance,wei2020multi} observed a split between \textit{one-to-one} and \textit{many-to-many} matching in visual-semantic embedding spaces characterized by the use of one or several embedding representations per image or caption. Song and Soleymani~\cite{song2019pvse} build many global representations for each image or sentence by using a multi-head self-attention on local descriptors. %
Other methods use region-level and word-level descriptors to build a global image-to-text similarity from many-to-many matching. Li \etal~\cite{Li2019VSRN} employ a graphical convolutional network~\cite{kipf2016semi} for semantic reasoning of region proposals obtained from a Faster-RCNN~\cite{ren2015faster} detector.
Veit \etal~\cite{veit2018separating} propose a conditional embedding approach to solve the multiplicity of hashtags, but it does not rely on a joint embedding space, hence cannot be %
directly applied to cross-modal retrieval.

Recently, the most successful way of addressing many-to-many image-to-sentence matching is through joint visual and textual reasoning modules appended on top of separate region-level encoders~\cite{lee2018stacked,liu2019focus,lu2019vilbert,12in1,nam2017dual,wang2019camp, wei2020multi,zhang2020context}. Most of such methods involve cross-modal attention networks and report state-of-the-art results on cross-modal retrieval. This, however, comes with a large increase in computational cost at test time: pairs formed by the query and every database entry need to go through the reasoning module. Focusing on scalability, we choose to build on top of approaches that directly utilize the joint embedding space and are compatible with large-scale indexing.

Finally, concurrent to our work, Wray~\etal~\cite{wray2021semantic} consider cross-modal video retrieval and discusses similar limitations of the one-to-one correspondence assumptions for evaluation. They propose to consider semantic similarity proxies computed on captions for a more reliable evaluation on standard video retrieval datasets.

\noindent\textbf{Probabilistic embedding.}
Probabilistic representations of data have a long history in machine learning~\cite{murphy2012machine}. They were introduced in 2014 for word embeddings~\cite{vilnis2014iclr}, as they gracefully handle the inherent hierarchies in language, since then, a line of research has explored different distribution families for word representations~\cite{li2018smoothing,neelakantan2014efficient,nguyen2017mixture}. Recently, probabilistic embeddings have been introduced for vision tasks. Oh~\etal~\cite{oh2019hibprobemb} proposed the Hedged Instance Embedding (HIB) to handle the one-to-many correspondences for metric learning, while other works apply probabilistic embeddings to face understanding~\cite{shi2019faceprobemb,chang2020datauncertainty}, 2D-to-3D pose estimation~\cite{sun2020viewinvprobemb}, speaker diarization~\cite{silnova2020probabilistic}, and prototype embeddings~\cite{scott2019icmlw}. Our work extends HIB to joint embeddings between images and captions, in order to represent the different levels of granularities in the two domains and to implicitly capture the resulting one-to-many associations. Recently Sch\"onnfeld \etal~\cite{schonfeld2019generalized} utilized Variational Autoencoders~\cite{Kingma2014vae} for zero-shot recognition. Their latent space is conceptually similar to ours, but is learned and used in very different ways: they simply use a 2-Wasserstein distance as their distribution alignment loss and learn classifiers on top, while \pcme uses a probabilistic \emph{contrastive} loss that enables us to use the latent features directly for retrieval. To our knowledge, \pcme is the first work that uses probabilistic embeddings for multi-modal retrieval.

\section{Method}
\label{sec:method}

In this section, we present our \textbf{\pcmefull} (\textbf{\pcme}) framework and discuss its conceptual workings and advantages.

We first define the cross-modal retrieval task. Let $\cD=(\cC,\cI)$ denote a vision and language dataset, where $\cI$ is a set of images and $\cC$ a set of captions. The two sets are connected via ground-truth matches. For a caption $c \in \cC$ (respectively an image $i\in\cI$), the set of corresponding images (respectively captions) is given by $\tau(c)\subseteq \cI$ (respectively $\tau(i)\subseteq\cC$). Note that for every query $q$, there may be multiple cross-modal matches ($|\tau(q)|>1$). Handling this multiplicity will be the central focus of our study.

Cross-modal retrieval methods typically learn an embedding space $\realnum^D$ such that we can quantify the subjective notion of ``similarity'' into the distance between two vectors. For this, two embedding functions $\visencoder$, $\textencoder$ are learned to map image and text samples into the common space $\realnum^D$. 

\subsection{Building blocks for \pcme}
\label{subsec:building-blocks}

We introduce two key ingredients for \pcme: joint visual-textual embeddings and probabilistic embeddings.

\begin{figure*}
    \centering
    \includegraphics[width=.9\linewidth]{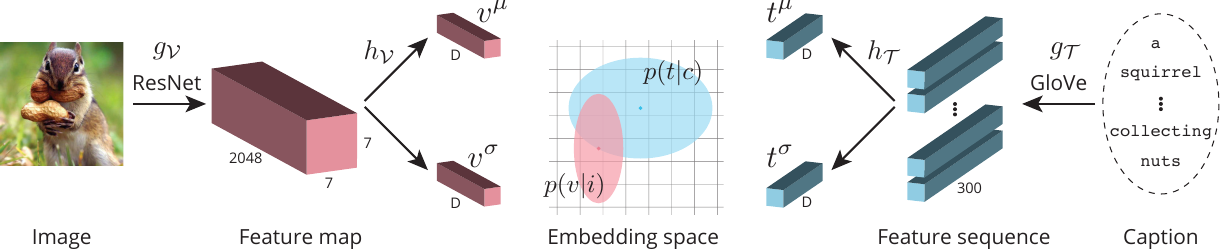}
    \vspace{.0em}
    \caption{\small\textbf{Method overview.} The visual and textual encoders for \pcmefull (\pcme) are shown. Each modality outputs mean and variance vectors in $\realnum^D$, which represent a normal distribution in $\realnum^D$.}
    \label{fig:method-overview}
    \vspace{-.5em}
\end{figure*}

\subsubsection{Joint visual-textual embeddings}
\label{subsubsec:joint-visual-textual-embeddings}

We describe how we learn visual and textual encoders. We then present a previous attempt at addressing the multiplicity of cross-modal associations.

\noindent\textbf{Visual encoder $\visencoder$.} 
We use the ResNet image encoder~\cite{he2016_cvpr_resnet}. Let $\vzf = \visfeat(i): \cI \rightarrow \realnum^{h \times w \times d_v} $ denote the output before the global average pooling (GAP) layer. Visual embedding is computed via $\vf = \vishead(\vzf)\in\realnum^D$ where in the simplest case $\vishead$ is the GAP followed by a linear layer. We modify $\vishead$ to let it predict a distribution, rather than a point.

\noindent\textbf{Textual encoder $\textencoder$.} 
Given a caption $c$, we build the array of word-level descriptors $\tzf = \textfeat(c)\in \realnum^{L(c)\times d_t}$, where $L(c)$ is the number of words in $c$. We use the pre-trained GloVe~\cite{pennington2014glove}. The sentence-level feature $t$ is given by a bidirectional GRU~\cite{cho2014properties}: $t=\texthead(\tzf)$ on top of the GloVe features.

\noindent\textbf{Losses used in prior work.}
The joint embeddings are often learned with a contrastive or triplet loss~\cite{faghri2018vsepp,frome2013devise}.

\noindent\textbf{Polysemous visual-semantic embeddings (PVSE)~\cite{song2019pvse}}
are designed to model one-to-many matches for cross-modal retrieval. PVSE adopts a multi-head attention block on top of the visual and textual features to encode $K$ possible embeddings per modality. For the visual case, each visual embedding $\vf^k\in \realnum^D$ for $k\in\{1,\ldots,K\}$ is given by: $\vf^k = \layernorm \left( \vishead(\vzf) + \sigmoid(\fcweight^1\att_\cV^k(\vzf)\vzf) \right)$, where $\fcweight^1 \in \realnum^{d_v \times D}$ are the weights of fully connected layers, $\sigmoid$ is the sigmoid function and $\layernorm$ is the LayerNorm~\cite{ba2016layer}. $\att_\cV^k$ denotes the $k$-th attention head of the visual self-attention $\att_\cV$. Textual embeddings $\tf^k$ for $k\in\{1,\ldots,K\}$ are given symmetrically by the multi-head attention: $\tf^k = \layernorm \left( \texthead(\tzf) + \sigmoid(\fcweight^2\att_\cC^k(\tzf)\tzf)\right)$. PVSE learns the visual and textual encoders with the multiple instance learning (MIL) objective, where only the best pair among the $K^2$ possible visual-textual embedding pairs is supervised.

\subsubsection{Probabilistic embeddings for a single modality}

Our \pcme models each sample as a distribution. It builds on the Hedged Instance Embeddings (HIB)~\cite{oh2019hibprobemb}, a single-modality methodology developed for representing instances as a distribution. HIB is the probabilistic analogue of the contrastive loss~\cite{hadsell2006dimensionality}. HIB trains a probabilistic mapping $p_\theta(z|x)$ that not only preserves the pairwise semantic similarities but also represents the inherent uncertainty in data. We describe the key components of HIB here.

\noindent\textbf{Soft contrastive loss.}
To train $p_\theta(z|x)$ to capture pairwise similarities, HIB formulates a soft version of the contrastive loss~\cite{hadsell2006dimensionality} widely used for training deep metric embeddings. For a pair of samples $(x_\alpha,x_\beta)$, the loss is defined as:
\begin{equation}
\label{eq:hib_soft_cont}
\mathcal{L}_{\alpha\beta}(\theta)=
\begin{cases}
-\log p_\theta(m| x_\alpha, x_\beta)&\text{if }\alpha,\beta\text{ is a match} \\
-\log \left(1-p_\theta(m|x_\alpha, x_\beta)\right)&\text{otherwise}
\end{cases}
\end{equation}
where $p_\theta(m|x_\alpha,x_\beta)$ is the \textit{match probability}.

\noindent\textbf{Factorizing match probability.} 
\cite{oh2019hibprobemb} has factorized $p_\theta(m|x_\alpha,x_\beta)$ into the match probability based on the embeddings $p(m|z_\alpha,z_\beta)$ and the encoders $p_\theta(z|x)$. This is done via Monte-Carlo estimation:
\begin{equation}
\label{eq:hib_match_probability_monte_carlo}  
p_\theta(m|x_\alpha, x_\beta) \approx \frac{1}{J^2}\sum_j^J \sum_{j^\prime}^J p(m|z_\alpha^{j}, z_\beta^{j^\prime})  
\end{equation}
where $z^{j}$ are samples from the embedding distribution $p_\theta(z|x)$. For the gradient to flow, the embedding distribution should be reparametrization-trick-friendly~\cite{kingma2013auto}. 

\noindent\textbf{Match probability from Euclidean distances.}
We compute the sample-wise match probability as follows:
\begin{equation}
\label{eq:hib_match_probability_from_embeddings}
p(m| z_\alpha, z_\beta) = \sigmoid (-a \| z_\alpha - z_\beta \|_2 + b)
\end{equation}
where $(a,b)$ are learnable scalars and $\sigmoid(\cdot)$ is sigmoid.

\subsection{Probabilistic cross-modal embedding (PCME)} 
\label{subsec:pcme}
We describe how we learn a joint embedding space that allows for probabilistic representation with \pcme.

\subsubsection{Model architecture}
\label{subsubsec:pcme-architecture}

\begin{figure}
    \centering
    \includegraphics[width=.9\linewidth]{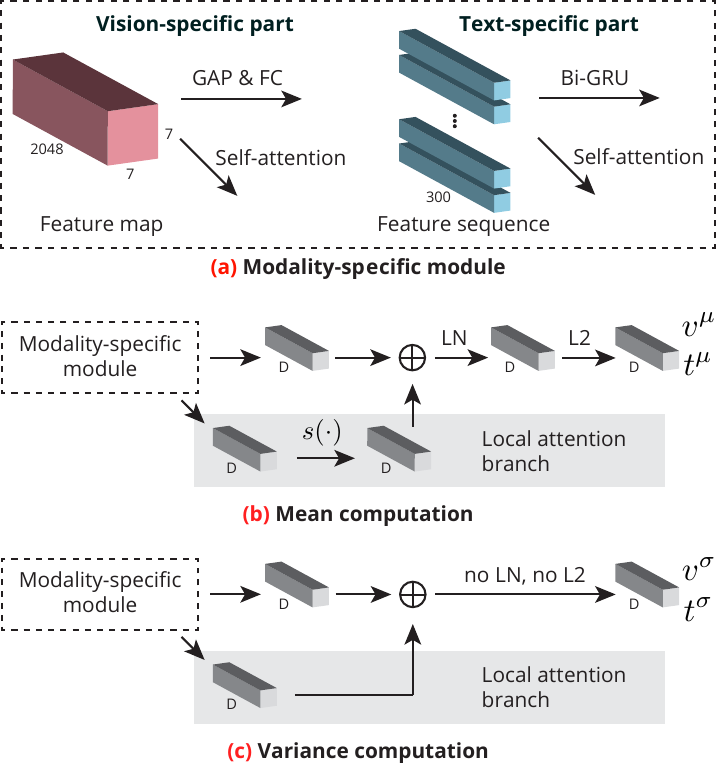}
    \caption{\small\textbf{Head modules.} The visual and textual heads ($\vishead$, $\texthead$) share the same structure, except for modality-specific modules \textcolor{red}{(a)}. The mean \textcolor{red}{(b)} and variance \textcolor{red}{(c)} computations differ: variance module does not involve sigmoid $s(\cdot)$, LayerNorm (LN), and L2 projection.
    }
    \label{fig:method-head-modules}
    \vspace{-.5em}
\end{figure}

An overview of \pcme is shown in Figure~\ref{fig:method-overview}. \pcme represents an image $i$ and caption $c$ as normal distributions, $p(\vf|i)$ and $p(\tf|c)$ respectively, over the same embedding space $\realnum^D$. We parametrize the normal distributions with mean vectors and diagonal covariance matrices in $\realnum^D$:
\begin{align}
\begin{split}
\label{eq:visual-textual-distributions}
    p(\vf|i)\sim N\left(\vishead^\mu(\vzf), \text{diag}(\vishead^\sigma(\vzf)\right)\\
    p(\tf|c)\sim N\left(\texthead^\mu(\tzf), \text{diag}(\texthead^\sigma(\tzf)\right)
\end{split}
\end{align}
where $\vzf=\visfeat(i)$ is the feature map and $\tzf=\textfeat(c)$ is the feature sequence (\S\ref{subsubsec:joint-visual-textual-embeddings}). For each modality, two head modules, $\generichead^\mu$ and $\generichead^\sigma$, compute the mean and variance vectors, respectively. They are described next.

\noindent\textbf{Local attention branch.} Inspired by the PVSE architecture (\S\ref{subsubsec:joint-visual-textual-embeddings}, \cite{song2019pvse}), we consider appending a \emph{local attention branch} in the head modules ($\generichead^\mu$, $\generichead^\sigma$) both for image and caption encoders. See Figure~\ref{fig:method-head-modules} for the specifics. The local attention branch consists of a self-attention based aggregation of spatial features, followed by a linear layer with a sigmoid activation function. We will show with ablative studies that the additional branch helps aggregating spatial features more effectively, leading to improved performance.

\noindent\textbf{Module for $\mu$ versus $\sigma$.}
Figure~\ref{fig:method-head-modules} shows the head modules $\generichead^\mu$ and $\generichead^\sigma$, respectively. For $\vishead^\mu$ and $\texthead^\mu$, we apply sigmoid in the local attention branch and add the residual output. In turn, LayerNorm (LN)~\cite{ba2016layer} and L2 projection operations are applied~\cite{song2019pvse,vaswani2017attention}. For $\vishead^\sigma$ and $\texthead^\sigma$, we observe that the sigmoid and LN operations overly restrict the representation, resulting in poor uncertainty estimations (discussed in \S\fref{D}).
We thus do not use sigmoid, LN, and L2 projection for the uncertainty modules.

\noindent\textbf{Soft cross-modal contrastive loss.}
Learning the joint probabilistic embedding is to learn the parameters for the mappings $p(\vf|i)=p_{\theta_v}(\vf|i)$ and $p(\tf|c)=p_{\theta_t}(\tf|c)$. We adopt the probabilistic embedding loss in Equation~\eqref{eq:hib_soft_cont}, where the match probabilities are now based on the cross-modal pairs $(i,c)$: $\mathcal{L}_{\text{emb}}(\theta_v,\theta_t;i,c)$, where $\theta=(\theta_v,\theta_t)$ are parameters for visual and textual encoders, respectively. The match probability is now defined upon the visual and textual features: $p_\theta(m|i, c) \approx \frac{1}{J^2}\sum_j^J \sum_{j^\prime}^J \sigmoid (-a \| \vf^j - \tf^{j^\prime} \|_2 + b) $ where $\vf^j$ and $\tf^{j^\prime}$ follow the distribution in Equation~\eqref{eq:visual-textual-distributions}. 

\noindent\textbf{Additional regularization techniques.}
We consider two additional loss functions to regularize the learned uncertainty. Following~\cite{oh2019hibprobemb}, we prevent the learned variances from collapsing to zero by introducing the KL divergence loss between the learned distributions and the standard normal $\mathcal{N} (0, I)$. We also employ the \emph{uniformity loss} that was recently introduced in ~\cite{wang2020understanding}, computed between all embeddings in the minibatch. See \S\fref{A.1} for more details.

\noindent\textbf{Sampling SGD mini-batch.}
We start by sampling $B$ ground-truth image-caption matching pairs $(i,c) \in \cG$. Within the sampled subset, we consider \textit{every} positive and negative pair dictated by the ground truth matches. This would amount to $B$ matching pairs and $B(B-1)$ non-matching pairs in our mini-batch.

\noindent\textbf{Measuring instance-wise uncertainty.}
The covariance matrix predicted for each input represents the inherent uncertainty for the data. For a scalar uncertainty measure, we take the determinant of the covariance matrix, or equivalently the geometric mean of the $\sigma$'s. Intuitively, this measures the volume of the distribution.

\subsubsection{How does our loss handle multiplicity, really?}

We perform a gradient analysis to study how our loss in Equation~\eqref{eq:hib_soft_cont} handles multiplicity in cross-modal matches and learn uncertainties in data. In \S\fref{A.2}, we further make connections with the MIL loss used by PVSE (\S\ref{subsubsec:joint-visual-textual-embeddings}, \cite{song2019pvse}).

We first define the distance logit: $l_{jj^\prime}:=-a \| \vf^j - \tf^{j^\prime} \|_2 + b$ and compare the amount of supervision with different $(j,j^\prime)$ values. To see this, take the gradient on $l_{jj^\prime}$.
\begin{align}
\label{eq:soft-contrastive-gradient}
\frac{\partial \mathcal L_{\text{emb}}}{\partial l_{jj^\prime}}& = 
\begin{cases}
w_{jj^\prime}\cdot(1 - s(l_{jj^\prime})) & \text{\small for positive match}\\
-w_{jj^\prime}\cdot s(l_{jj^\prime}) & \text{\small for negative match}
\end{cases}
\\
w_{jj^\prime}&:=\frac{e^{\pm l_{jj^\prime}}}{\sum_{\alpha \alpha^\prime}e^{\pm l_{\alpha \alpha^\prime}}}\quad\text{\small where $\pm$ is the positivity of match}.\nonumber
\end{align}
We first observe that if $w_{jj^\prime}=1$, then Equation~\eqref{eq:soft-contrastive-gradient} is exactly the supervision from the soft contrastive loss (Equation~\eqref{eq:hib_soft_cont}). Thus, it is the term $w_{jj^\prime}$ that let the model learn multiplicity and represent associated uncertainty.

To study the behavior of $w_{jj^\prime}$, first assume that $(v,t)$ is a positive pair. Then, $w_{jj^\prime}$ is the softmax over the pairwise logits $l_{jj^\prime}$. Thus, pairs with smaller distances $\|\vf^j-\tf^{j^\prime}\|_2$ have greater weights $w_{jj^\prime}$ than distant ones. Similarly, if $(v,t)$ is negative pair, then $w_{jj^\prime}$ assigns greater weights on distant pairs than close ones. In other words, $w_{jj^\prime}$ gives more weights on pair samples that correctly predicts the distance relationships on the embedding space. This results in a reward structure where wrong similarity predictions do not get penalized significantly, as long as there is at least one correct similarity prediction. Such a reward encourages the embeddings to produce more diverse samples and hedge the bets through non-zero values of $\sigma$ predictions.

\subsubsection{Test-time variants}
\label{subsec:test-time-alternative}

Unlike methods that employ cross-modal reasoning modules~\cite{lee2018stacked,liu2019focus,lu2019vilbert,12in1,nam2017dual,wang2019camp,wei2020multi,zhang2020context}, computing match probabilities at test time for \pcme reduces to computing a function over pairwise Euclidean distances. This means that the probabilistic embeddings of \pcme can be used in various ways for computing the match probabilities at test time, with different variants having different computational complexities. The options are split into two groups. 
\begin{enumerate*}
    \item \noindent\textbf{Sampling-based variants.} Similar to training, one can use Monte-Carlo sampling (Equation~\eqref{eq:hib_match_probability_monte_carlo}) to approximate match probabilities. Assuming $J$ samples, this requires $O(J^2)$ distance computations per match, as well as $O(J^2)$ space for every database entry. This implies that $J$ plays an important role in terms of test time complexity.
    \item \noindent\textbf{Non-sampling variants.} One can simply use the distances based on $\mu$ to approximate match probabilities. In this case, both time and space complexities become $O(1)$. We ablate this variant (``$\mu$ only") in our experiments, as it is directly comparable to deterministic approaches. We also may use any distributional distance measures with closed-form expressions for Gaussian distributions. Examples include the 2-Wasserstein distance, Jensen Shanon (JS) divergence, and Expected Likelihood Kernel (ELK). We ablate them as well. The details of each probabilistic distance can be found in \S\fref{B}.
\end{enumerate*}

\section{Experiments}
\label{sec:exp}

We present experimental results for \pcme. We start with the experimental protocol and a discussion on the problems with current cross-modal retrieval benchmarks and evaluation metrics, followed by alternative solutions (\S\ref{subsec:protocol}). We then report experimental results on the CUB cross-modal retrieval task (\S\ref{subsec:pcme-variants}) and COCO (\S\ref{subsec:coco-results}). We present an analysis of the embedding space in \S\ref{subsec:uncertainty}.

\subsection{Experimental protocol}
\label{subsec:protocol}

\begin{figure}
    \centering
    \small
    \includegraphics[width=.92\columnwidth]{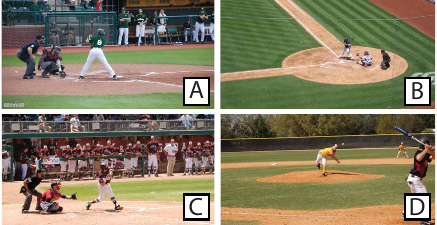} \\
    \vspace{.5em}
    \resizebox{\columnwidth}{!} {
    \begin{tabular}{l}
         a) A baseball player swinging a bat at a ball.  \\
         b) A baseball player is getting ready to hit a ball. \\
         c) A baseball player standing next to home plate holding a bat. \\
         d) A group of baseball players at the pitch. \\
    \end{tabular}
    }
    \vspace{.5em}
    \caption{Can you match the captions to the images? In the COCO annotations, each of the four captions corresponds to (only) one of the four images (Answer: \raisebox{\depth}{\rotatebox{180}{A:b, B:c, C:a, D:d}}).}
    \label{fig:coco-caption-ambiguity}
    \vspace{-.5em}
\end{figure}

We use ResNet~\cite{he2016_cvpr_resnet} pre-trained on ImageNet and the pre-trained GloVe with 2.2M vocabulary~\cite{pennington2014glove} for initializing the visual and textual encoders. Training proceeds in two phases: a warm-up phase where only the head modules are trained, followed by end-to-end fine-tuning of all parameters. We use a ResNet-152 (resp. ResNet-50) backbone with embedding dimension $D=1024$ (resp. $D=512$) for MS-COCO (resp. CUB). For both datasets, models are always trained with Cutout~\cite{devries2017cutout} and random caption dropping~\cite{bowman-etal-2016-generating} augmentation strategies with 0.2 and 0.1 erasing ratios, respectively. We use the AdamP optimizer~\cite{heo2021adamp} with the cosine learning rate scheduler~\cite{loshchilov2016sgdr} for stable training. More implementation details are provided in \S\fref{C.2}.
Hyperparameter details and ablations are presented in \S\fref{D}. 

\subsubsection{Metrics for cross-modal retrieval}
\label{subsubsec:metrics}

Researchers have long been aware of many potentially positive matches in the cross-modal retrieval evaluation sets. They use metrics that reflect such consideration. 

Many works report the \textbf{Recall@$k$} (R@$k$) metrics with varying numbers for $k$. This evaluation policy, with larger values of $k$, becomes more lenient to plausible wrong predictions prevalent in COCO. However, it achieves leniency at the cost of failing to penalize obviously wrong retrieved samples. The lack of penalties for wrongly retrieved top-$k$ samples may be complemented by the precision metrics.

Musgrave \etal~\cite{musgrave2020metric} proposed the \textbf{R-Precision} (R-P) metric as an alternative; for every query $q$, we compute the ratio of positive items in the top-$r$ retrieved items, where $r=|\tau(q)|$ is the number of ground-truth matches. This precision metric has a desirable property that a retrieval model achieves the perfect R-Precision score if and only if it retrieves all the positive items before the negatives.

For R-Precision to make sense, all the existing positive pairs in a dataset must be annotated. Hence, we expand the existing ground truth matches by seeking further plausible positive matches in a database through extra information (\eg class labels for COCO). More concretely, a pair $(i,c)$ is declared positive if the binary label vectors for the two instances, $y^i,y^c\in \{0,1\}^{d_{label}}$, differ at most at $\zeta$ positions. In practice, we consider multiple criteria $\zeta \in \{ 0, 1, 2\}$ and average the results with those $\zeta$ values. We refer to metrics based on such class-based similarity as \textbf{Plausible Match (PM)} because we incentivize models to retrieve plausible items. We refer to the R-Precision metric based on the Plausible Match policy as \textbf{PMRP}. More details in \S\fref{C.1}.

\subsubsection{Cross-modal retrieval benchmarks}
\label{subsubsec:benchmarks}

\textbf{COCO Captions}~\cite{chen2015microsoft} is a widely-used dataset for cross-modal retrieval models. It consists of 123,287 images from MS-COCO~\cite{lin2014microsoft} with 5 human-annotated captions per image. We present experimental results on COCO. We follow the evaluation protocol of~\cite{karpathy2015deep} where the COCO validation set is added to the training pool (referred to as rV or rVal in~\cite{BeansBurgers,faghri2018vsepp}). Our training and validation splits contain 113,287 and 5,000 images, respectively. We report results on both 5K and (the average over 5-fold) 1K test sets.

The problem with COCO as a cross-modal retrieval benchmark is the binary relevance assignment of image-caption pairs $(i,c)$. As a result, the number of matching captions $\tau(i)$ for an image $i$ is always 5. Conversely, the number of matching images $\tau(c)$ for a caption $c$ is always 1. All other pairs are considered non-matching, independent of semantic similarity. This is far from representing the semantic richness of the dataset. See Figure~\ref{fig:coco-caption-ambiguity} for an illustration. While all $4\times 4$ possible pairs are plausible positive pairs, 12 of them are assigned negative labels during training and evaluation. This results in noisy training and, more seriously, unreliable evaluation results. 

We re-purpose the CUB 200-2011~\cite{welinder2010caltech} as a more reliable surrogate for evaluating cross-modal retrieval models. We utilize the caption annotations by Reed \etal~\cite{reed2016learning}; they consist of ten captions per image on CUB images (11,788 images of 200 fine-grained bird categories). False positives are suppressed by the fact that the captions and images are largely homogeneous within a class. False negatives are unlikely to happen because the images contain different types of birds across classes and the captions are generated under the instruction that the annotators should focus on class-distinguishing characteristics~\cite{reed2016learning}.

We follow the class splits proposed by Xian \etal~\cite{xian2017zero}, where 150 classes are used for training and validation, and the remaining 50 classes are used for the test. The hyperparameters are validated on the 150 training classes. We refer to this benchmark as \textit{CUB Captions}.

\subsection{Results on CUB}
\label{subsec:pcme-variants}

\paragraph{Similarity measures for retrieval at test time.}
We have discussed alternative similarity metrics that \pcme may adopt at test time (\S~\ref{subsec:test-time-alternative}). The ``Mean only'' metric only uses the $\generichead^\mu$ features, as in deterministic retrieval scenarios. It only requires $O(N)$ space to store the database features. Probabilistic distance measures like ELK, JS-divergence, and 2-Wasserstein, require the storage for $\mu$ and $\sigma$ features, resulting in the doubled storage requirement. Sampling-based distance computations, such as the average L2 distance and match probability, need $J^2$ times the storage required by the Mean-only baseline.

We compare the above variants in Table~\ref{tab:retrieval-strateiges} and \S\fref{E.1}.
First of all, we observe that \pcme, with any test-time similarity measure, mostly improves over the deterministically trained \pcme ($\mu$-only training). Even if the test-time similarity is computed as if the embeddings are deterministic (Mean only), \pcme training improves the retrieval performances (24.7\% to 26.1\% for i2t and 25.6\% to 26.7\% for t2i). Other cheaper variants of probabilistic distances, such as 2-Wasserstein, also result in reasonable performances (26.2\% and 26.7\% for i2t and t2i, respectively), while introducing only twice the original space consumption. The best performance is indeed attained by the similarity measure using the match probability, with 26.3\% and 26.8\% i2t and t2i performances, respectively. There exists a trade-off between computational cost and performance and the deterministic test-time similarity measures. We use the match probability measure at test time for the rest of the paper.

\paragraph{Comparison against other methods.}
We compare \pcme against VSE0~\cite{faghri2018vsepp} and PVSE~\cite{song2019pvse} in Table~\ref{tab:cub-comparison}. As an important ingredient for PVSE, we consider the use of the hardest negative mining (HNM). We first observe that PVSE with HNM tends to obtain better performances than VSE0 under the R@1 metric, with 47.8\% for $K$=4, compared to 44.2\% for VSE0. However, under the R-Precision metric, we observe all PVSE models with HNM are worse than VSE0 (R-Precision drops from 22.4\% for VSE0 to 18.4\% for PVSE $K$=4). It seems that PVSE with HNM tends to retrieve items based on diversity, rather than precision. We conjecture that the HNM is designed to optimize the R@1 performances; more details in \S\fref{E.2}.
Comparing PVSE with different values of $K$, we note that increasing $K$ does not always bring about performance gains under the R-Precision metric (20.5\%, 21.2\% and 19.9\% for $K$=1,2,4, respectively, for t2i), while the improvement is more pronounced under the R@1 metric. Finally, \pcme provides the best performances on both R-Precision and R@1 metrics, except for the R@1 score for i2t. \pcme also improves upon its deterministic version, \pcme $\mu$-only, with some margin: +1.6 pp and +1.2 pp on i2t and t2i R-Precision scores, respectively.

\begin{table}[t]
    \centering
    \small
    \resizebox{\columnwidth}{!} {
    \setlength{\tabcolsep}{3pt}
    \begin{tabular}{l|c|l|c|c|c}
    \toprule
        \multirow{2}{*}{\shortstack{\pcme\\variant}} & \multirow{2}{*}{Sampling} & \multirow{2}{*}{\shortstack{Test-time \\ Similarity Metric}} & \multirow{2}{*}{\shortstack{Space \\ complexity}} & \multirow{2}{*}{\shortstack{i2t \\ R-P}} & \multirow{2}{*}{\shortstack{t2i \\ R-P}} \\ &&&& \\ \midrule
        $\mu$ only & \xmark & Mean only & $O(N)$ & 24.70 & 25.64 \\ \midrule
        \multirow{6}{*}{\pcme} & \xmark & Mean only & $O(N)$ & 26.14 & 26.67 \\
         & \xmark & ELK & $O(2N)$ & 25.33 & 25.87 \\
         & \xmark & JS-divergence & $O(2N)$ & 25.06 & 25.55 \\
         & \xmark & 2-Wasserstein & $O(2N)$ & \underline{26.16} & \underline{26.69} \\
         & \cmark & Average L2 & $O(J^2 N)$ & 26.11 & 26.64 \\
         & \cmark & Match prob & $O(J^2 N)$ & \textbf{26.28} & \textbf{26.77} \\ \bottomrule
    \end{tabular}
    }
    \vspace{.5em}
    \caption{\small\textbf{Pairwise distances for distributions.} There are many options for computing the distance between two distributions. What are the space complexity and retrieval performances for each option? R-P stands for the R-Precision.
    }
    \label{tab:retrieval-strateiges}
    \vspace{-.5em}
\vspace{1em}
\small
\centering
\setlength{\tabcolsep}{.86em}
\begin{tabular}{@{}l|c|cc|cc@{}}
\toprule
\multirow{2}{*}{Method} & \multirow{2}{*}{HNM} & \multicolumn{2}{c}{Image-to-text} & \multicolumn{2}{|c}{Text-to-image}            \\
& & R-P & R@1 &  R-P & R@1 \\ \midrule
VSE0      & \xmark      & 22.4       & 44.2    &  22.6        & 32.7    \\ \midrule
PVSE K=1 & \cmark      & 22.3       & 40.9    &  20.5        & 31.7    \\
PVSE K=2 & \cmark      & 19.7       & 47.3    &  21.2       & 28.0    \\
PVSE K=4 & \cmark      & 18.4       & \textbf{47.8}    &  19.9       & 34.4    \\ \midrule
PCME $\mu$ only   & \xmark      & 24.7       & 46.4    &  25.6   & \textbf{35.5}    \\
PCME       & \xmark      & \textbf{26.3}       & 46.9    &  \textbf{26.8}       & 35.2   \\ \bottomrule
\end{tabular}
\vspace{.5em}
\caption{\small {\bf Comparison on CUB Caption test split.} R-P and R@1 stand for R-Precision and Recall@1, respectively. The usage of hardest negative mining (HNM) is indicated.}
\label{tab:cub-comparison}
\vspace{-.5em}
\end{table}

\subsection{Results on COCO}
\label{subsec:coco-results}

As we have identified potential problems with measuring performance on COCO (\S\ref{subsubsec:benchmarks}), we report the results with our Plausible-Match R-Precision (PMRP) metrics (\S\ref{subsubsec:metrics}) that captures the model performances more accurately than the widely-used R@$k$ metrics. Table~\ref{tab:coco} shows the results with state-of-the-art COCO retrieval methods. We observe that the stochastic version of \pcme performs better than the deterministic variant ($\mu$ only) across the board. In terms of the R@1 metric, PVSE $K$=2~\cite{song2019pvse}, VSRN~\cite{Li2019VSRN} and AOQ~\cite{chen2020adaptive} work better than \pcme (\eg 45.2\%, 53.0\%, 55.1\% versus 44.2\% for the 5K, i2t task). However, on the more accurate PMRP metric, \pcme outperforms previous methods with some margin (\eg 31.8\%, 29.7\%, 33.0\% versus 34.1\% for the 5K, i2t task). The results on two metrics imply that \pcme retrieves the plausible matches much better than previous methods do. The full results can be found in \S\fref{E}.

\begin{table}[t]
\centering
\small
\resizebox{\columnwidth}{!} {
\setlength{\tabcolsep}{3pt}
\begin{tabular}{@{}l|cc|cc|cc|cc@{}}
\toprule
&\multicolumn{4}{c}{1K Test Images} & \multicolumn{4}{|c}{5K Test Images} \\ \midrule
\multirow{2}{*}{Method} & \multicolumn{2}{|c}{i2t} & \multicolumn{2}{|c}{t2i} & \multicolumn{2}{|c}{i2t} & \multicolumn{2}{|c}{t2i}  \\
& {\footnotesize PMRP} & {\footnotesize R@1} & {\footnotesize PMRP} & {\footnotesize R@1} & {\footnotesize PMRP} & {\footnotesize R@1} & {\footnotesize PMRP}  & {\footnotesize R@1} \\ \midrule
VSE++~\cite{faghri2018vsepp} & -  & 64.6  & -  & 52.0 & - & 41.3 & - & 30.3 \\
PVSE K=1~\cite{song2019pvse} & 40.3$^*$  & 66.7 & 41.8$^*$ & 53.5 & 29.3$^*$ & 41.7 & 30.1$^*$ & 30.6 \\
PVSE K=2~\cite{song2019pvse} & 42.8$^*$ & 69.2 & 43.6$^*$ & 55.2 & 31.8$^*$ & 45.2 & 32.0$^*$ & 32.4 \\ 
VSRN~\cite{Li2019VSRN} & 41.2$^*$ & 76.2 & 42.4$^*$ & 62.8 & 29.7$^*$ & 53.0 & 29.9$^*$ & 40.5
 \\ 
VSRN + AOQ~\cite{chen2020adaptive} & 44.7$^*$ & \textbf{77.5} & 45.6$^*$ & \textbf{63.5} & 33.0$^*$ & \textbf{55.1} & 33.5$^*$ & \textbf{41.1} \\\midrule
\pcme {\footnotesize $\mu$ only} & \textbf{45.0}{\color{white} *} & 68.0 & 45.9{\color{white} *} & 54.6 & 34.0{\color{white} *} & 43.5 & 34.3{\color{white} *} & 31.7\\
\pcme & \textbf{45.0}{\color{white} *} & 68.8 & \textbf{46.0}{\color{white} *} & 54.6 & \textbf{34.1}{\color{white} *} & 44.2 & \textbf{34.4}{\color{white} *} & 31.9 \\ \bottomrule
\end{tabular}
}
\vspace{.5em}
\caption{\small {\bf Comparison on MS-COCO.} 
PMRP stands for the Plausible Match R-Precision and R@1 for Recall@1. ``$*$'' denotes results produced by the published models.}
\label{tab:coco}
\vspace{-.5em}
\end{table}

\begin{figure*}
    \centering
    \includegraphics[width=\linewidth]{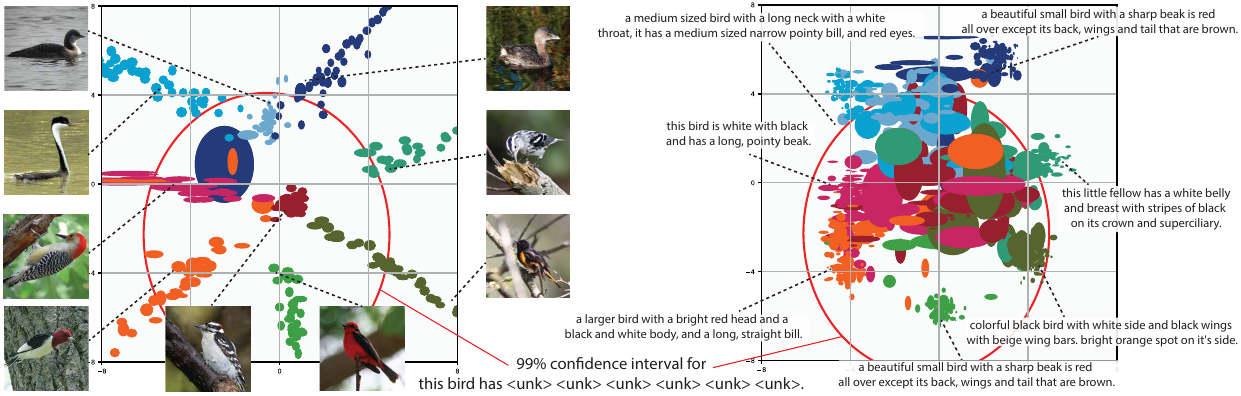}
    \caption{\small\textbf{Visualization of the probabilistic embedding.} The learned image (left) and caption (right) embeddings on 9 subclass of CUB Captions. Classes are color-coded. Each ellipse shows the 50\% confidence region for each embedding. The red ellipse corresponds to the generic CUB caption, ``this bird has $<$unk$>$ $\cdots$ $<$unk$>$'' with 99\% confidence region.
    }
    \label{fig:cub_2d_toy}
    \vspace{-.75em}
\end{figure*}

\begin{figure}
    \centering
    \includegraphics[width=0.5\columnwidth]{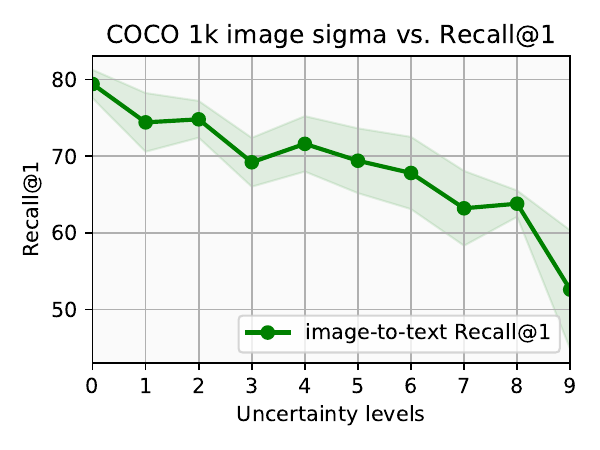}%
    \includegraphics[width=0.5\columnwidth]{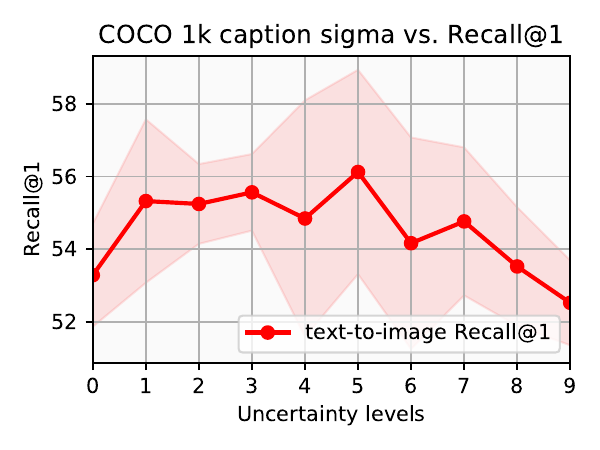}
    \caption{\small\textbf{$\sigma$ versus performance.} Performance of \pcme at different per-query uncertainty levels in COCO 1k test set.
    }
    \label{fig:uncertainty-vs-performance}
    \vspace{-.5em}
\end{figure}

\begin{figure}
    \centering
    \includegraphics[width=.5\columnwidth]{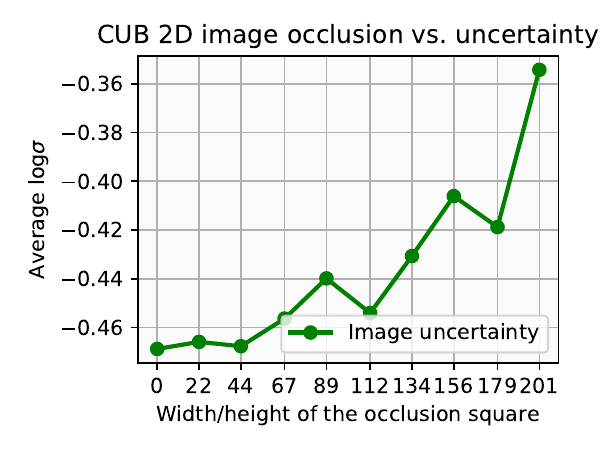}%
    \includegraphics[width=.5\columnwidth]{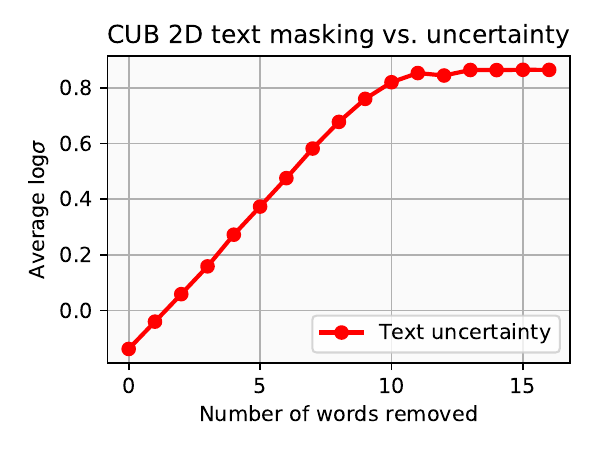}
    \caption{\small\textbf{$\sigma$ captures ambiguity.} Average $\sigma$ values at different ratios of erased pixels (for images) and words (for captions).
    }
    \label{fig:uncertainty-vs-erasing}
    \vspace{-.75em}
\end{figure}

\subsection{Understanding the learned uncertainty}
\label{subsec:uncertainty}

Having verified the retrieval performance of \pcme, we now study the benefits of using probabilistic distributions for representing data. We show that the learned embeddings not only represent the inherent uncertainty of data but also enable set algebras among samples that roughly correspond to their semantic meanings.

\noindent\textbf{Measuring uncertainty with $\sigma$.} 
In an automated decision process, it benefits a lot to be able to represent uncertainty. For example, the algorithm may refrain from making a decision based on the uncertainty estimates. We show that the learned cross-modal embeddings capture the inherent uncertainty in the instance. We measure the instance-wise uncertainty for all query instances by taking the geometric mean over the $\sigma\in\realnum^D$ entries (\S\ref{subsubsec:pcme-architecture}). We then compute the average R@1 performances in each of the 10 uncertainty bins. Figure~\ref{fig:uncertainty-vs-performance} plots the correlation between the uncertainty and R@1 on the COCO test set. We observe performance drops with increasing uncertainty. In \S\fref{F.2},
we visualize which word affects more to uncertainty. Example uncertain instances and their retrieval results are in \S\fref{F.3}.

\noindent\textbf{2D visualization of \pcme.}
To visually analyze the behavior of \pcme, we conduct a 2D toy experiment by using 9 classes of the CUB Captions (details in \S\fref{C.3}).
Figure~\ref{fig:cub_2d_toy} visualizes the learned image and caption embeddings. We also plot the embedding for the most generic caption for the CUB Captions dataset, ``this bird has $<$unk$>$ $<$unk$>$ \ldots'', where $<$unk$>$ is a special token denoting the absence of a word.  This generic caption covers most of the caption variations in the embedding space (red ellipses).

\begin{figure}
    \centering
    \includegraphics[width=\columnwidth]{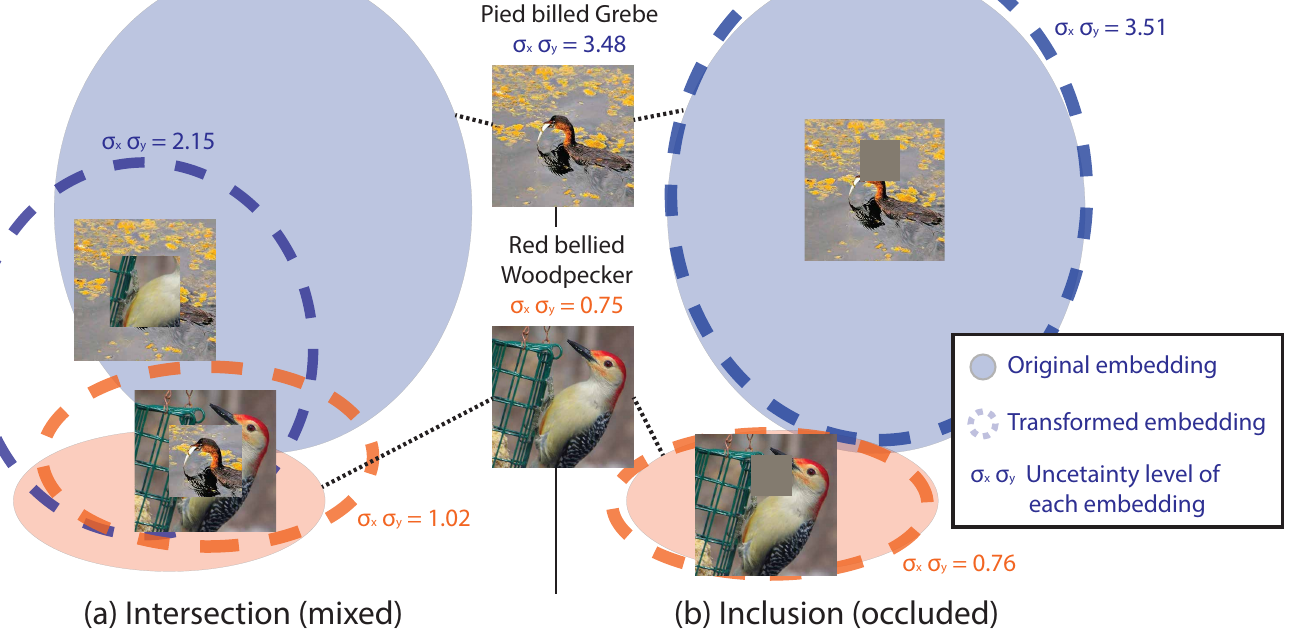}  
    \caption{\small\textbf{Set algebras.}
    For two images, we visualize the embeddings for either erased or mixed samples. Mixing (left) and erasing (right) operations roughly translate to the intersection and inclusion relations between the corresponding embeddings.}
    \label{fig:set-algebras}
    \vspace{-.75em}
\end{figure}

\noindent\textbf{Set algebras.}
To understand the relationship among distributions on the embedding space, we artificially introduce different types of uncertainties on the image data. In Figure~\ref{fig:set-algebras}, we start from two bird images and perform erasing and mixing transformations~\cite{yun2019cutmix}. On the embedding space, we find that the mixing operation on the images results in embeddings that cover the \textit{intersection} of the original embeddings. Occluding a small region in input images, on the other hand, amounts to slightly wider distributions, indicating an \textit{inclusion} relationship. We quantitatively verify that the sigma values positively correlate with the ratio of erased pixels in Figure~\ref{fig:uncertainty-vs-erasing}. In COCO, we observe a similar behavior (shown in \S\fref{F.1}).
We discover another positive correlation between the caption ambiguity induced by erasing words and the embedding uncertainty.

\section{Conclusion}

We introduce \pcmefull (\pcme) that learns probabilistic representations of multi-modal data in the embedding space.
The probabilistic framework provides a powerful tool to model the widespread one-to-many associations in image-caption pairs.
To our knowledge, this is the first work that uses probabilistic embeddings for a multi-modal task. 
We extensively ablate our \pcme and show that not only it improves the retrieval performance over its deterministic counterpart, but also provides uncertainty estimates that render the embeddings more interpretable.

\section*{Acknowledgements}
We thank our NAVER AI Lab colleagues for valuable discussions. All experiments were conducted on NAVER Smart Machine Learning (NSML)~\cite{nsml} platform.

{\small
\bibliographystyle{ieee_fullname}
\bibliography{egbib}
}

\newcommand{\hpref}[1]{\textcolor{red}{#1\xspace}}
\newcommand{\ve}[1]{\boldsymbol{#1\xspace}}

\appendix
\numberwithin{equation}{section}
\numberwithin{figure}{section}
\numberwithin{table}{section}

\section*{Supplementary Materials}
We include additional materials in this document. We describe additional details on \pcme to complement the main paper (\S\ref{sec:supp-pcme-detail}). Various probabilistic distances are introduced (\S\ref{sec:supp-probdist}). We provide the experimental protocol details (\S\ref{sec:supp-protocol}), ablation studies (\S\ref{sec:supp-ablation}), and additional results (\S\ref{sec:supp-moreresults}). Finally, more uncertainty analyses are shown (\S\ref{sec:supp-moreuncertainty}).

\section{More details for \pcme}
\label{sec:supp-pcme-detail}

In this section, we provide details for \pcme.

\subsection{The uniformity loss}
\label{sec:supp-uniformity}
Recently, Wang~\etal~\cite{wang2020understanding} proposed the uniformity loss which enforces the feature vectors to distribute uniformly on the unit hypersphere. In Wang~\etal~\cite{wang2020understanding}, the uniformity loss was shown to lead to better representations for L2 normalized features. Since our $\mu$ vectors are projected to the unit L2 hypersphere, we also employ the uniformity loss to learn better representations. We apply the uniformity loss on the joint embeddings $\mathcal Z = \{v_1^1, t_1^1, \ldots, v_B^J, t_B^J\}$ in the mini-batch size of $B$ as follows:
\begin{equation}
\label{eq:pcme_uniform}
    \mathcal L_{\text{Unif}} = \sum_{z, z^\prime \in \mathcal Z \times \mathcal Z} e^{-2 \| z - z^\prime \|_2^2}.
\end{equation}

\subsection{Connection between the soft contrastive loss and the MIL objective of PVSE}
\label{sec:supp-softcont-mil}

In the main text, we presented an analysis based on gradients to study how the loss function in %
Equation~(\textcolor{red2}{1})
handles plurality in cross-modal matches and learns uncertainties in data. Here we make connections with the MIL loss used by PVSE 
(\S\textcolor{red2}{3.1.1}, 
\cite{song2019pvse}); this section follows the corresponding section in the main paper.

To build connections with PVSE, consider a one-hot weight array $w_{jj^\prime}$ where, given that $(v,t)$ is a positive pair, the ``one'' value is taken only by the single pair $(j,j^\prime)$ whose distance is smallest. Define $w_{jj^\prime}$ for a negative pair $(v,t)$ conversely. Then, we recover the MIL loss used in PVSE, where only the best match among $J^2$ predictions are utilized. As we see in the experiments, our \emph{softmax} weight scheme provides more interpretable and performant supervision for the uncertainty than the \emph{argmax} version used by PVSE.

\section{Probabilistic distances}
\label{sec:supp-probdist}
We introduce probabilistic distance variants to measure the distance between two normal distributions $p = \mathcal N (\mu_1, \sigma_1^2)$ and $q = \mathcal N (\mu_2, \sigma_2^2)$. All distance functions are non-negative and become zero if and only if two distributions are identical. Extension to multivariate Gaussian distributions with diagonal variance can be simply derived by taking the summation over the dimension-wise distances.

\textbf{Kullback–Leibler (KL) divergence} measures the difference between two distributions as follows:
\begin{align}
\label{eq:supp-kl-divergence}
\begin{split}
    KL(p, q) &= \int \log \frac{p}{q} dp \\
    &= \frac{1}{2} \left[ \log \frac{\sigma_2^2}{\sigma_1^2} + \frac{\sigma_1^2}{\sigma_2^2} + \frac{(\mu_1 - \mu_2)^2}{\sigma_2^2} \right].
\end{split}
\end{align}
KL divergence is not a metric because it is asymmetric ($KL(p, q) \neq KL (q, p)$) and does not satisfy the triangular inequality. If $q$ has a very small variance, nearly zero, the KL divergence between $p$ and $q$ will be explored. In other words, if we have a very certain embedding, which has nearly zero variance, in our gallery set, then the certain embedding will be hardly retrieved by KL divergence measure. In the latter section, we will show that KL divergence leads to bad retrieval performances in the real-world scenario.

\textbf{Jensen-Shannon (JS) divergence} is the average of forward ($KL(p, q)$) and reverse ($KL(q, p)$) KL divergences. Unlike KL divergence, the square root of JS divergence is a metric function.
\begin{equation}
\label{eq:supp-js-divergence}
    JS(p, q) = \frac{1}{2} \left [KL(p, q) + KL(q, p) \right].
\end{equation}
Like KL divergence, JS divergence still has division term by variances $\sigma_1, \sigma_2$, it can be numerically unstable when the variances are very small.

\textbf{Probability product kernels}~\cite{jebara2004probability} are generalized inner product for two distributions, that is:
\begin{equation}
\label{eq:supp-ppk}
    PPK(p, q) = \int p(z)^\rho q(z)^\rho dz.
\end{equation}
When $\rho = 1$, it is called the expected likelihood kernel (ELK), and when $\rho = 1/2$, it is called Bhattacharyya’s affinity~\cite{bhattacharyya1943measure}, or Bhattacharyya kernel.

\textbf{Expected likelihood kernel (ELK)} is a special case of PPK when $\rho=1$ in Equation \eqref{eq:supp-ppk}. In practice, we take log to compute ELK as follows:
\begin{equation}
\label{eq:supp-elk}
    ELK(p, q) = \frac{1}{2}\left[ \frac{(\mu_1 - \mu_2)^2}{\sigma_1^2 + \sigma_2^2} + \log (\sigma_1^2 + \sigma_2^2) \right].
\end{equation}

\textbf{Bhattacharyya kernel (BK)} is another special case of PPK when $\rho=1/2$ in Equation \eqref{eq:supp-ppk}. The log BK is defined as follows:
\begin{equation}
\label{eq:supp-bk}
    BK(p, q) = \frac{1}{4}\left[ \frac{(\mu_1 - \mu_2)^2}{\sigma_1^2 + \sigma_2^2} + 2 \log (\frac{\sigma_2}{\sigma_1} + \frac{\sigma_1}{\sigma_2}) \right].
\end{equation}

\textbf{Wasserstein distance} is a metric function of two distributions on a given metric space $M$. The Wasserstein distance between two normal distributions on $\mathbb R^1$, 2-Wasserstein distance, is defined as follows:
\begin{equation}
\label{eq:supp-wasserstein}
    W(p, q)^2 = (\mu_1 - \mu_2)^2 + {\sigma_1 - \sigma_2}^2.
\end{equation}

\section{Experimental Protocol Details}
\label{sec:supp-protocol}

We introduce the cross-modal retrieval benchmarks considered in this work. We discuss the issues with the current practice for the evaluation and introduce new alternatives.

\subsection{Plausible Match R-Precision (PMRP) details}
\label{sec:supp-pmrp-details}
In this work, we seek more reliable sources of pairwise similarity measurements through class and attribute labels on images. For example, on the CUB caption dataset, we have established the positivity of pairs by the criterion that a pair $(i,c)$ is positive if and only if both elements in the pair belong to the same bird class. Similarly, on the COCO caption dataset, we judge the positivity through the multiple class labels (80 classes total) attached per image: a pair $(i,c)$ is positive if and only if the binary class vectors for the two instances, $y^i,y^c\in \{0,1\}^{80}$, differ at most at $\zeta$ positions (Hamming Distance). In MS-COCO 5k test images, 48 images do not have instance labels; we omit them during the evaluation.
Note that because we use R-Precision, the ratio of positive items in top-$r$ retrieved items where $r$ is the number of the ground-truth matches, increasing $\zeta$ will make $r$ larger, and will penalize methods more, which retrieve irrelevant items.

\begin{figure}
    \centering
    \includegraphics[width=\linewidth]{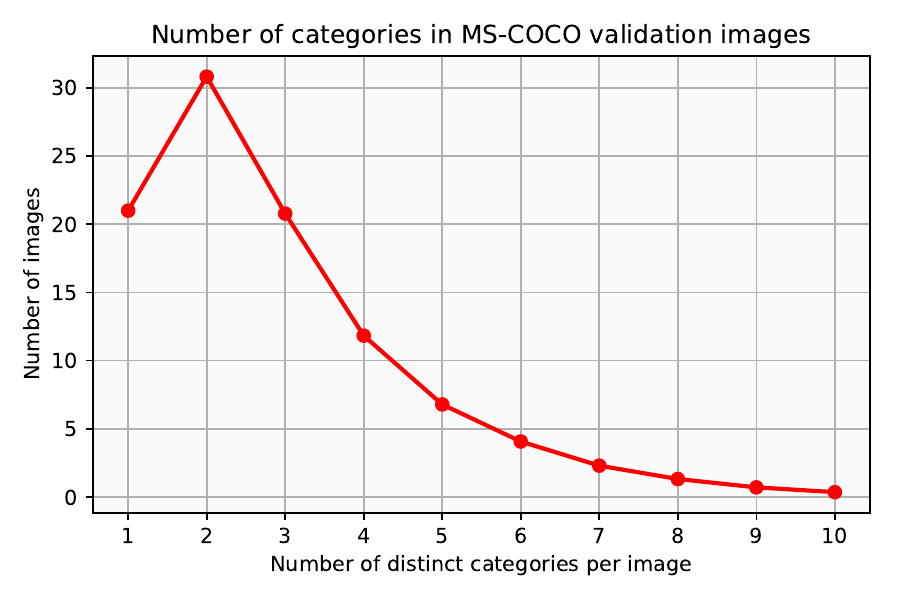}
    \caption{\small\textbf{Number of distinct categories in MS-COCO validation set.} Images that have more than 10 categories are omitted.}
    \label{fig:supp_num_categories_coco}
\end{figure}

\begin{figure*}
    \centering
    \includegraphics[width=\linewidth]{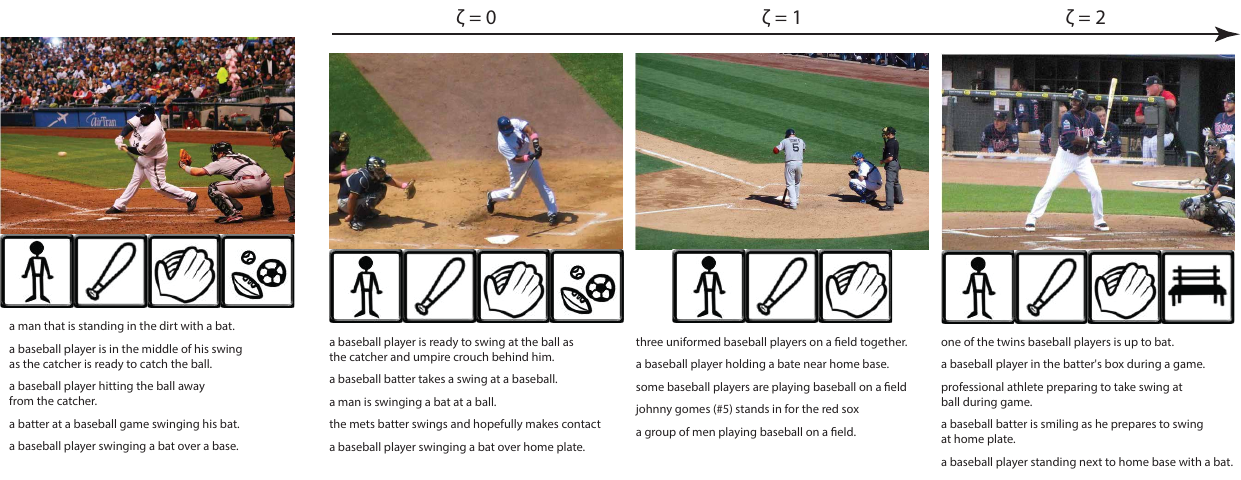}
    \caption{\small\textbf{MS-COCO plausible match examples.} The plausible examples of the most left instance from $\zeta=0$ to $\zeta=2$. The contained instance classes, $\zeta$, figure and captions are shown.}
    \label{fig:supp_pmrp}
\end{figure*}

In Figure~\ref{fig:supp_num_categories_coco}, we visualize the number of distinct categories per image in the MS-COCO validation set. In the figure, we can observe that about the half of the images have more than two categories. To avoid penalty caused by almost neglectable objects (as shown in Figure~\ref{fig:supp_pmrp}), we set $\zeta = 2$ for measuring the PMRP score. For PMRP with different $\zeta$ rather than 2, results can be found in \S\ref{sec:supp-moreresults}.

\subsection{Implementation details}
\label{sec:supp-implementation-details}

\paragraph{Common.}
As in Faghri \etal~\cite{faghri2018vsepp}, we use ResNet~\cite{he2016_cvpr_resnet} pre-trained on ImageNet and the pre-trained GloVe with 2.2M vocabulary~\cite{pennington2014glove} for initializing the visual and textual encoders ($\visencoder$, $\textencoder$). We first warm-up the models by training the head modules for each modality, with frozen feature extractors. Afterwards, the whole parameters are fine-tuned in an end-to-end fashion. We use the ResNet-152 backbone with embedding dimension $D=1024$ for MS-COCO and ResNet-50 with $D=512$ for CUB. For all experiments, we set the number of samples $J=7$ (the detailed study is in \S\ref{sec:supp-moreresults}). We use AdamP optimizer~\cite{heo2021adamp} with the cosine learning rate scheduler~\cite{loshchilov2016sgdr} for stable training.

\paragraph{MS-COCO.}
We follow the evaluation protocol of~\cite{karpathy2015deep} where the validation set is added to the training pool (referred to as rV in~\cite{BeansBurgers,faghri2018vsepp}). Our training and validation splits contain 113,287 and 5,000 images, respectively. We report results on both 5K and (the average over 5-fold) 1K test sets.

\paragraph{Hyperparameter search protocol.}
We validate the initial learning rate, number of epochs for the warm-up and fine-tuning, and other hyperparameters on the 150 CUB training classes and the MS-COCO caption validation split. For MS-COCO, we use the initial learning rate as 0.0002, 30 warm-up and 30 finetune epochs. Weights for regularizers $\mathcal L_\text{KL}$ and $\mathcal L_\text{Unif}$ are set to 0.00001 and 0, respectively. For CUB Caption, the initial learning rate is 0.0001, the number of warm-up epochs 10 and fine-tuning epochs 50. Weights for regularizers $\mathcal L_\text{KL}$ and $\mathcal L_\text{Unif}$ are set to 0.001 and 10, respectively. For both datasets, models are always trained with Cutout~\cite{devries2017cutout} and random caption dropping~\cite{bowman-etal-2016-generating} augmentation strategies with 0.2 and 0.1 erasing ratios, respectively. The initial values for $a, b$ in 
Equation (\textcolor{red2}{3})
are set to -15 and 15 for COCO (-5 and 5 for CUB), respectively.

\subsection{CUB 2D toy experiment details}
\label{sec:supp-cub-toy}
We select nine bird classes from CUB caption; three swimming birds (``Western Grebe'', ``Pied Billed Grebe'', ``Pacific Loon''), three small birds (``Vermilion Flycatcher'', ``Black And White Warbler'', ``American Redstart''), and three woodpeckers (``Red Headed Woodpecker'', ``Red Bellied Woodpecker'', ``Downy Woodpecker'').

We slightly modify \pcme to learn 2-dimensional embeddings. For the image encoder, we use the same structure as the other experiments, but omitting the attention modules from the $\mu$ and $\sigma$ modules. For the caption encoder, we train 1024-dimensional bi-GRU on top of GloVe vectors and apply two 2D projections to get the 1024 dimensional $\mu$ and $\sigma$ embedding. The other training details are the same as the other CUB caption experiments.

\section{Ablation studies}
\label{sec:supp-ablation}

We provide ablation studies on \pcme for regularization terms, $\sigma$ module architectures, the number of samples $J$ during training, and embedding dimension $D$.

\paragraph{Regularizing uncertainty.}
\pcme predicts probabilistic outputs. We have considered uncertainty-specific regularization strategy in the main paper, the information bottleneck loss $\mathcal L_\text{KL}$ and the uniform loss $\mathcal L_\text{Unif}$. We study the benefits of those ingredients. Table~\ref{tab:uncertainty-regularization} shows our results. We report cross-validated MAP@R~\cite{musgrave2020metric} on the 150 class training CUB caption datasets. The KL loss increases the sigma values to a meaningful range (from $e^{-13.01}\approx 2.2\times 10^{-6}$ to $e^{-3.84}\approx 0.02$. The uniformity loss prevents the uncertainty from collapsing and slightly improves performances. 

\begin{table}
    \small
    \centering
    \begin{tabular}{@{}c|c|c|c|c|c@{}}
    \toprule
        &     & i2t       & t2i      & Image & Caption \\
    $\mathcal L_\text{KL}$ & $\mathcal L_\text{Unif}$ & MAP@R & MAP@R & $\mathbb E [\log \sigma]$ & $\mathbb E [\log \sigma]$ \\ \midrule
    \xmark   & \xmark       & 10.56     & 13.32     & -13.01      & -8.77         \\
    \cmark   &  \xmark      & 10.57     & 13.77     & -3.84       & -3.89         \\
    \xmark   & \cmark       & 10.56     & 13.31     & -11.26      & -7.59         \\
    \cmark   & \cmark       & \textbf{10.65}     & \textbf{13.84}     & -3.63       & -3.64         \\ \bottomrule
    \end{tabular}
    \vspace{.5em}
    \caption{\small\textbf{Regularization for uncertainty.} Cross-validated MAP@R performances on CUB training set, with and without KL and uniformity loss terms. The scale estimate $\mathbb{E}[\log\sigma]$ is an averaged value over the $\sigma$ dimensions as well as the validation samples.}
    \label{tab:uncertainty-regularization}
\end{table}

\begin{table}
\small
\centering
\begin{tabular}{@{}lccc@{}}
\toprule
Method            & \hspace{-1em}DoF($\sigma$) & i2t & t2i \\ \midrule
PCME {\footnotesize $\mu$ only} &  0  & 24.7         & 25.6         \\
PCME {\footnotesize isotropic} &  1   & 25.7         & 26.0         \\
PCME             & 512  & \textbf{26.3}         & \textbf{26.8}         \\ \bottomrule
\end{tabular}
\vspace{.5em}
\caption{\small\textbf{DoF for $\sigma$.} R-Precision on the CUB Caption test set.}
\label{tab:sigma-isotropic}
\end{table}

\begin{figure}
    \centering
    \includegraphics[width=\columnwidth]{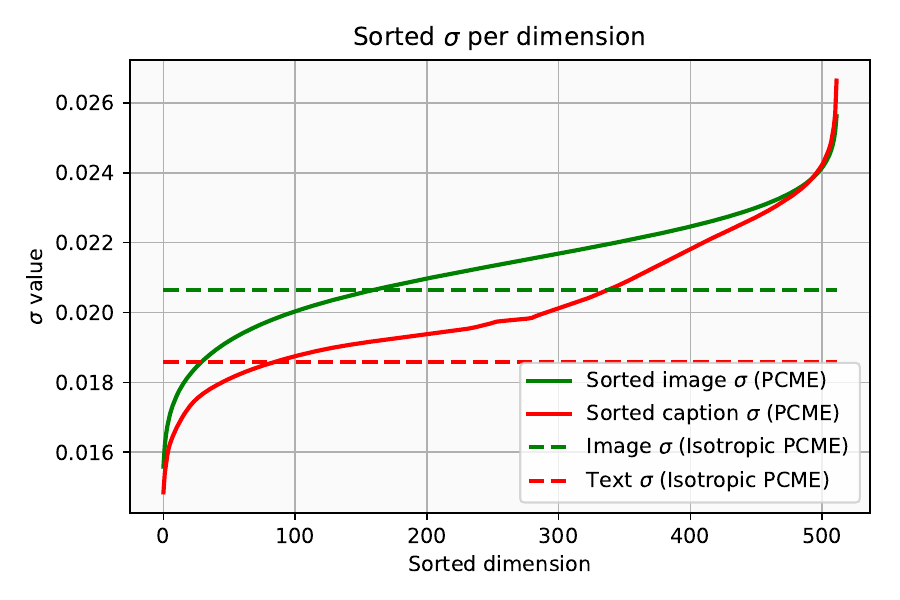}
    \caption{\small\textbf{How isotropic are variances?} Sorted values of variance are compared against the trained values of isotropic \pcme. Results on CUB test set.}
    \label{fig:sigma-isotropic}
\end{figure}

\paragraph{DoF for $\sigma$.}
Though by default we parametrize the full diagonal elements of the covariance matrix $\Sigma\in\realnum^{D\times D}$ with the vector $\sigma\in\realnum^D$, one may parametrize $\sigma$ more cheaply via \eg a scalar, by restricting the embedding distribution family to isotropic Gaussians. Table~\ref{tab:sigma-isotropic} shows the trade-off between the degree of freedom (DoF) for $\sigma$ and the R-Precision of \pcme. Indeed, allowing greater degrees of freedom for $\sigma$ brings better performance. Figure~\ref{fig:sigma-isotropic} shows the average variance values for each dimension, which supports that the learned variances require high DoF.

\paragraph{Architecture study.}

\begin{table}[t]
    \small
    \centering
    \resizebox{\columnwidth}{!} {
    \begin{tabular}{c|c|c|c}
\toprule
$\mu$ & $\sigma$ & I-to-T & T-to-I \\
local attention & local attention & R-Precision & R-Precision \\ \midrule
\xmark & \xmark & 25.60 & 25.85 \\
\xmark & \cmark & 24.65 & 25.15 \\
\cmark & \xmark & 25.01 & 25.52 \\
\cmark & \cmark & \textbf{26.28} & \textbf{26.77} \\ \bottomrule
\end{tabular}
}
\begin{tabular}{c}
~
\end{tabular}
\resizebox{\columnwidth}{!} {
\begin{tabular}{@{}c|c|c@{}}
\toprule
$s(\cdot)$ \& $\texttt{LN}$ in $\sigma$ module & I-to-T R-Precision & T-to-I R-Precision \\ \midrule
\cmark & 23.81   & 24.58   \\
\xmark & \textbf{26.28}   & \textbf{26.77}   \\ \bottomrule
\end{tabular}
}
\vspace{.5em}
\caption{\small\textbf{Architectures for $\mu$ and $\sigma$.} Architecture design choices comparison on CUB caption test split.}
\label{tab:cub-arch}
\end{table}

Table~\ref{tab:cub-arch} shows the architecture design comparisons for \pcme on CUB Caption test split. In the table, applying local attention to both $\mu$ and $\sigma$ modules performs the best. Furthermore, we ablate sigmoid and $\texttt{LN}$ parts of $\sigma$ modules, which can restrict the representation of variances. As a result, limiting representations by sigmoid and layer norm harms the final performances.

\paragraph{Number of samples during training.}

\begin{figure}
    \centering
    \includegraphics[width=.9\columnwidth]{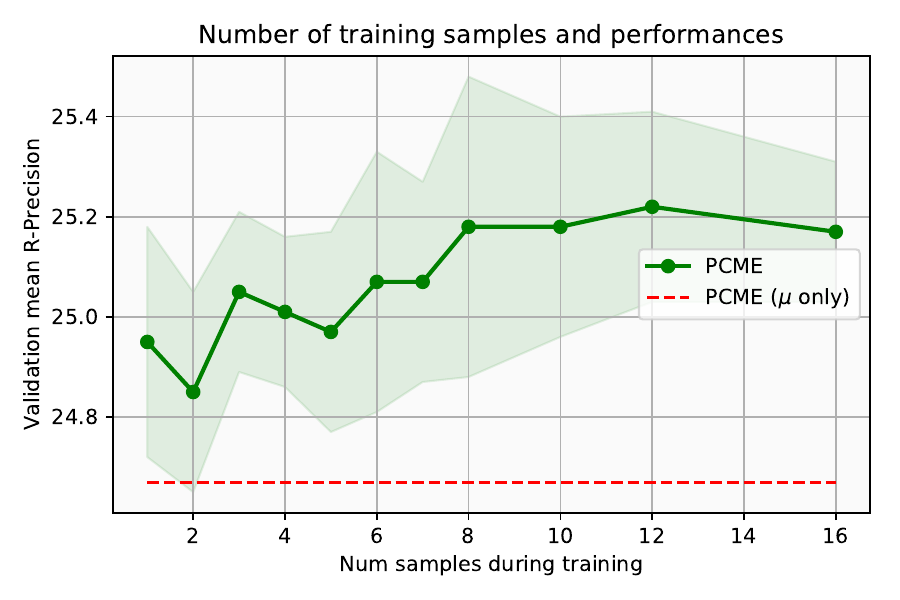}
    \caption{\small\textbf{Number of samples.} The cross-validated \pcme performances against the number of samples $J$ during training.}
    \label{fig:number-of-samples}
\end{figure}

In Figure~\ref{fig:number-of-samples}, we report the cross-validated mean R-Precision scores by varying the number of samples $J$ during training. In the figure, we observe that larger $J$ leads to higher performances. In practice, we choose $J=7$ for computation budgets.

\begin{figure}
    \includegraphics[width=.9\columnwidth]{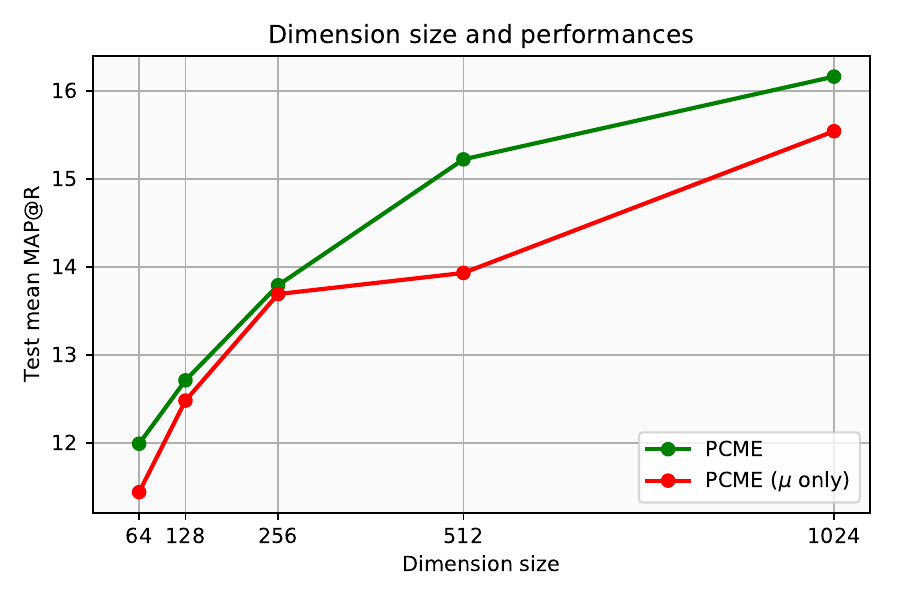}
    \caption{\small\textbf{Embedding dimensions.} The \pcme performance against the embedding dimensions $D$.}
    \label{fig:embedding-dimensions}
\end{figure}

\paragraph{Embedding dimensions.}
Performances against different embedding space dimensions for \pcme $\mu$ only and \pcme are illustrated in Figure~\ref{fig:embedding-dimensions}. In all embedding dimensions, our stochastic approach (\pcme) consistently outperforms the deterministic approach (\pcme $\mu$ only).

\section{More results}
\label{sec:supp-moreresults}

In this section, we provide additional experimental results for \pcme on CUB Caption and COCO Caption.

\subsection{More results on similarity measures for retrieval at test time}
\label{sec:supp-more-testtime-variants}
\begin{table}[]
    \centering
    \small
    \resizebox{\columnwidth}{!} {
    \setlength{\tabcolsep}{3pt}
    \begin{tabular}{l|c|l|c|c|c}
    \toprule
        \multirow{2}{*}{\shortstack{\pcme\\variant}} & \multirow{2}{*}{Sampling} & \multirow{2}{*}{\shortstack{Test-time \\ Similarity Metric}} & \multirow{2}{*}{\shortstack{Space \\ complexity}} & \multirow{2}{*}{\shortstack{i2t \\ R-P}} & \multirow{2}{*}{\shortstack{t2i \\ R-P}} \\ &&&& \\ \midrule
        $\mu$ only & \xmark & Mean only & $O(N)$ & 24.70 & 25.64 \\ \midrule
        \multirow{6}{*}{\pcme} & \xmark & Mean only & $O(N)$ & 26.14 & 26.67 \\
         & \xmark & KL-divergence & $O(2N)$ & 21.99 & 20.92 \\
         & \xmark & JS-divergence & $O(2N)$ & 25.06 & 25.55 \\
         & \xmark & ELK & $O(2N)$ & 25.33 & 25.87 \\
         & \xmark & Bhattacharyya & $O(2N)$ & 24.93 & 25.27 \\
         & \xmark & 2-Wasserstein & $O(2N)$ & \underline{26.16} & \underline{26.69} \\
         & \cmark & Average L2 & $O(J^2 N)$ & 26.11 & 26.64 \\
         & \cmark & Match prob & $O(J^2 N)$ & \textbf{26.28} & \textbf{26.77} \\ \bottomrule
    \end{tabular}
    }
    \vspace{.5em}
    \caption{\small\textbf{Pairwise distances for distributions.} There are many options for computing the distance between two distributions. What are the space complexity and retrieval performances for each option? R-P stands for the R-Precision.
    }
    \label{tab:supp-retrieval-strateiges-full}
\end{table}

\begin{figure}
    \includegraphics[width=.9\columnwidth]{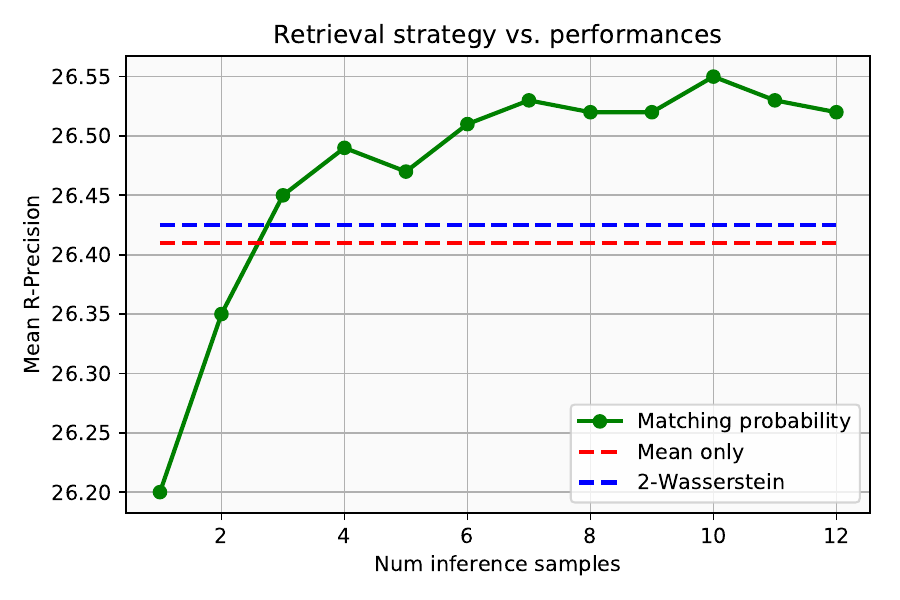}
    \caption{\small\textbf{Comparison of different retrieval strategies.}}
    \label{fig:retrieval-strategies}
\end{figure}

In Table~\ref{tab:supp-retrieval-strateiges-full}, we report the full retrieval results obtained by the different distribution distances discussed in \S\ref{sec:supp-probdist}. As discussed in \S\ref{sec:supp-probdist}, KL-divergence even shows worse results than the ``Mean only'' baseline, a non-probabilistic distance. We also report the performances against the number of samples of matching probability in Figure~\ref{fig:retrieval-strategies}. In the figure, the matching probability strategy shows better results than non-sampling strategies from $J=3$, and larger $J$ leads to better performances. Due to the computation complexity, we use $J=7$ in Table~\ref{tab:supp-retrieval-strateiges-full}.

\subsection{Discussion on hardest negative mining}
\label{sec:supp-hnm-discussion}

\begin{figure}
    \centering
    \includegraphics[width=\linewidth]{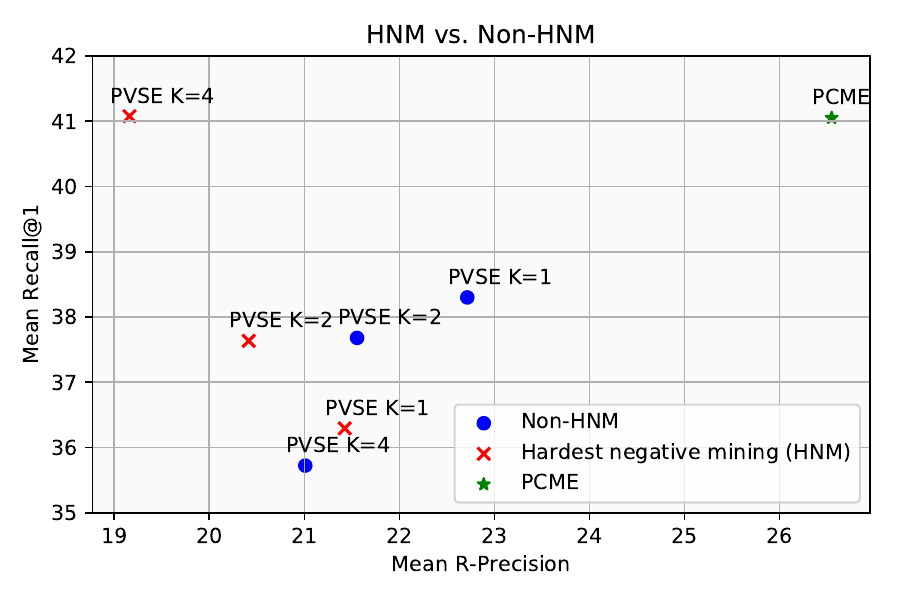}
    \caption{\small\textbf{Hardest negative mining (HNM) vs. Non-HNM.}}
    \label{fig:hmm_study}
\end{figure}

Since Recall@K is widely used for the evaluation of many cross-modal retrieval tasks, many recent cross-modal retrieval methods optimize Recall@1 directly by the hardest negative mining (HNM) strategy~\cite{faghri2018vsepp}, that is:
\begin{align}
\label{eq:hardest_negative_mining}
\begin{split}
    &\max_{t^\prime} [\alpha + \text{sim}(v, t^\prime) - \text{sim}(v, t) ] \\
    + &\max_{v^\prime} [\alpha + \text{sim}(v^\prime, t) - \text{sim}(v, t) ],
\end{split}
\end{align}
where $\text{sim}$ is the cosine similarity. This strategy neglects all other possible positive candidates, but only considers the most similar positive and negative pairs. To reveal that HM strategy disadvantages to learn the global structure, we measure two metrics on CUB caption, R-Precision and Recall@1. For non-HM strategy, we replace $\max$ to $\sum$ in Equation~\eqref{eq:hardest_negative_mining}. Figure~\ref{fig:hmm_study} shows R-Precision and recall@1 performances with different mining strategies. In the figure, PVSE with HNM strategy shows higher Recall@1 by increasing the number of embeddings $K$ (36.3 $\rightarrow$ 37.6 $\rightarrow$ 41.1), but at the same time, it reduces the R-Precision scores (21.4 $\rightarrow$ 20.4 $\rightarrow$ 19.2). On the other hand, for all $K$, Non-HNM strategy PVSE results show worse R@1 than HNM results but achieves higher R-Precision performances. 
In Table~\textcolor{red2}{3},
we show that this phenomenon is also observed in MS-COCO by measuring PMRP scores.

\subsection{Full results for CUB and COCO}

\begin{table}[]
\small
\centering
\setlength{\tabcolsep}{.86em}
\begin{tabular}{@{}l|c|cc|cc@{}}
\toprule
\multirow{2}{*}{Method} & \multirow{2}{*}{HNM} & \multicolumn{2}{c}{Image-to-text} & \multicolumn{2}{|c}{Text-to-image}            \\
& & R-P & R@1 &  R-P & R@1 \\ \midrule
VSE0      & \xmark      & 22.35       & 44.19    &  22.57        & 32.71    \\ \midrule
PVSE K=1 & \xmark      & 22.65       & 43.11    &  22.78        & 33.49    \\
PVSE K=2 & \xmark      & 21.62       & 44.05    &  21.49       & 31.31    \\
PVSE K=4 & \xmark      & 21.12       & 40.51    &  20.90       & 30.94    \\ \midrule
PVSE K=1 & \cmark      & 22.34       & 40.88    &  20.51        & 31.71    \\
PVSE K=2 & \cmark      & 19.67       & 47.29    &  21.16       & 27.98    \\
PVSE K=4 & \cmark      & 18.38       & \textbf{47.76}    &  19.94       & 34.39    \\ \midrule
PCME $\mu$ only   & \xmark      & 24.70       & 46.38    &  25.64   & \textbf{35.50}    \\
PCME       & \xmark      & \textbf{26.28}       & 46.92    &  \textbf{26.77}       & 35.22   \\ \bottomrule
\end{tabular}
\vspace{.5em}
\caption{\small {\bf Comparison on CUB Caption unseen 50 class test set.} R-P and R@1 stand for R-Precision and Recall@1, respectively. The usage of the hardest negative mining (HNM) is indicated.}
\label{tab:supp-cub-comparison-full}
\end{table}

\begin{table}[]
\small
\centering
\setlength{\tabcolsep}{.86em}
\begin{tabular}{@{}l|c|cc|cc@{}}
\toprule
\multirow{2}{*}{Method} & \multirow{2}{*}{HNM} & \multicolumn{2}{c}{Image-to-text} & \multicolumn{2}{|c}{Text-to-image}            \\
& & R-P & R@1 &  R-P & R@1 \\ \midrule
VSE0    & \xmark      & 19.85 & 40.88 & 18.72 & 25.51 \\ \midrule
PVSE K=1 & \xmark      & 19.69 & 40.65 & 18.72 & 25.58 \\
PVSE K=2 & \xmark      & 18.84 & 41.45 & 17.72 & 24.99 \\
PVSE K=4 & \xmark      & 18.31 & 38.08 & 17.21 & 23.54 \\ \midrule
PVSE K=1 & \cmark      & 18.98 & 38.77 & 18.23 & 23.49 \\
PVSE K=2 & \cmark      & 17.62 & 44.24 & 17.71 & 22.78 \\
PVSE K=4 & \cmark      & 17.47 & \textbf{44.98} & 17.44 & 26.19 \\ \midrule
PCME $\mu$ only & \xmark & 20.65 & 42.70 & 20.16 & \textbf{26.94} \\
PCME       & \xmark    & \textbf{20.87} & 43.10 & \textbf{20.37} & 26.47 \\ \bottomrule
\end{tabular}
\vspace{.5em}
\caption{\small {\bf Comparison on CUB Caption seen 150 class test set.} R-P and R@1 stand for R-Precision and Recall@1, respectively. The usage of the hardest negative mining (HNM) is indicated.}
\label{tab:supp-cub-comparison-full-seen}
\end{table}

\paragraph{CUB Caption.}
We report the full results on CUB Caption test data for unseen 50 classes and seen 150 classes in Table~\ref{tab:supp-cub-comparison-full} and Table~\ref{tab:supp-cub-comparison-full-seen}, respectively. In both splits, \pcme shows the best R-Precision performances against baselines.

\begin{table*}[t]
\centering
\small
\begin{tabular}{@{}l|c|cccc|cccc@{}}
\toprule
\multirow{2}{*}{Method} & \multirow{2}{*}{$D$}& \multicolumn{4}{c}{Image-to-text}   & \multicolumn{4}{|c}{Text-to-image}   \\
           && PMRP & R@1  & R@5  & R@10 & PMRP & R@1  & R@5  & R@10 \\ \midrule
VSE++~{\scriptsize BMVC'18}~\cite{faghri2018vsepp}    & 1024  & -     & 64.6 & 90.0 & 95.7 & -     & 52.0 & 84.3 & 92.0 \\
PVSE K=1~{\scriptsize CVPR'19}~\cite{song2019pvse}    & 1024 & 40.3$^*$  & 66.7 & 91.0 & 96.2 & 41.9$^*$  & 53.5 & 85.1 & 92.7 \\
PVSE K=2~{\scriptsize CVPR'19}~\cite{song2019pvse}    & 1024 $\times$ 2 & 42.8$^*$  & 69.2 & 91.6 & 96.6 & 43.7$^*$  & 55.2 & 86.5 & 93.7 \\ \midrule
PVSE K=4~{\scriptsize CVPR'19}~\cite{song2019pvse}    & 1024 $\times$ 4 & 41.5{\color{white} *} & 68.0 & 91.9 & 96.6 & 42.7{\color{white} *} & 54.1 & 85.5 & 92.9 \\
PVSE K=1 + SHM~\cite{schroff2015facenet}    & 1024 $\times$ 1 & 41.6{\color{white} *} & 66.1 & 91.4 & 96.4 & 42.4{\color{white} *} & 53.6 & 85.5 & 93.0 \\
PVSE K=2 + SHM~\cite{schroff2015facenet}    & 1024 $\times$ 2 & 39.0{\color{white} *} & 65.1 & 90.9 & 96.5 & 39.4{\color{white} *} & 53.1 & 85.4 & 93.0 \\ \midrule
VSRN~{\scriptsize ICCV'19}~\cite{Li2019VSRN} & 2048 & 41.2$^*$ & 76.2 & 94.8 & 98.2& 42.4$^*$ & 62.8 & 89.7 & 95.1 \\ 
VSRN + AOQ~{\scriptsize ECCV'20}~\cite{chen2020adaptive} & 2048 $\times$ 2 & 44.7$^*$ & \textbf{77.5} & \textbf{95.5} & \textbf{98.6} & 45.6$^*$ & \textbf{63.5} & \textbf{90.5} & \textbf{95.8} \\ \midrule
\pcme {\footnotesize $\mu$ only}    & 1024 & 45.0{\color{white} *}  & 68.0 & 92.0 & 96.2& 45.9{\color{white} *}  & 54.6 & 86.3 & 93.8 \\
\pcme       & 1024  $\times$ 2  & \textbf{45.1}{\color{white} *}  & 68.8 & 91.6 & 96.7 & \textbf{46.0}{\color{white} *}  & 54.6 & 86.3 & 93.8 \\ \bottomrule
\end{tabular}
\vspace{.5em}
\caption{\small {\bf 1K MS-COCO results.}
Plausible Match R-Precision (PMRP), Recall@K results on MS-COCO 1k test images. ``$*$'' denotes results produced by the published models.}
\label{tab:supp-coco-full-1k}
\end{table*}

\begin{table*}[t]
\centering
\small
\begin{tabular}{@{}l|c|cccc|cccc@{}}
\toprule
\multirow{2}{*}{Method} & \multirow{2}{*}{$D$}& \multicolumn{4}{c}{Image-to-text}   & \multicolumn{4}{|c}{Text-to-image}   \\
           && PMRP & R@1  & R@5  & R@10 & PMRP & R@1  & R@5  & R@10 \\ \midrule
VSE++~{\scriptsize BMVC'18}~\cite{faghri2018vsepp}    & 1024 & -    & 41.3 & 71.1 & 81.2 & -    & 30.3 & 59.4 & 72.4 \\
PVSE K=1~{\scriptsize CVPR'19}~\cite{song2019pvse}    & 1024 & 29.3$^*$ & 41.7 & 73.0 & 83.0 & 30.1$^*$ & 30.6 & 61.4 & 73.6 \\
PVSE K=2~{\scriptsize CVPR'19}~\cite{song2019pvse}    & 1024 $\times$ 2 & 31.8$^*$ & 45.2 & 74.3 & 84.5 & 32.0$^*$ & 32.4 & 63.0 & 75.0 \\ \midrule
PVSE K=4~{\scriptsize CVPR'19}~\cite{song2019pvse}    & 1024 $\times$ 4 & 30.5{\color{white} *} & 43.0 & 72.8 & 83.6 & 31.0{\color{white} *} & 31.2 & 61.5 & 74.4 \\
PVSE K=1 + SHM~\cite{schroff2015facenet}    & 1024 $\times$ 1 & 30.6{\color{white} *} & 41.1 & 71.6 & 82.7 & 30.8{\color{white} *} & 30.9 & 60.8 & 73.7 \\
PVSE K=2 + SHM~\cite{schroff2015facenet}    & 1024 $\times$ 2 & 28.1{\color{white} *} & 40.7 & 70.8 & 81.9 & 27.8{\color{white} *} & 29.9 & 60.4 & 73.4 \\ \midrule
VSRN~{\scriptsize ICCV'19}~\cite{Li2019VSRN} & 2048 & 29.7$^*$ & 53.0 & 81.1 & 89.4 & 29.9$^*$ & 40.5 & 70.6 & 81.1 \\ 
VSRN + AOQ~{\scriptsize ECCV'20}~\cite{chen2020adaptive} & 2048 $\times$ 2 & 33.0$^*$ & \textbf{55.1} & \textbf{83.3} & \textbf{90.8} & 33.5$^*$ & \textbf{41.1} & \textbf{71.5} & \textbf{82.0} \\ \midrule
\pcme {\footnotesize $\mu$ only} & 1024 & 34.0{\color{white} *} & 43.5 & 73.1 & 84.2 & 34.3{\color{white} *} & 31.7 & 62.2 & 74.9 \\
\pcme & 1024  $\times$ 2 & \textbf{34.1}{\color{white} *} & 44.2 & 73.8 & 83.6 & \textbf{34.4}{\color{white} *} & 31.9 & 62.1 & 74.5 \\ \bottomrule
\end{tabular}
\vspace{.5em}
\caption{\small {\bf Comparison on 5K MS-COCO.} 
Plausible Match R-Precision (PMRP), Recall@K results on MS-COCO 5k test images. ``$*$'' denotes results produced by the published models.}
\label{tab:supp-coco-full-5k}
\end{table*}

\paragraph{COCO Caption.}
We report the full results on MS-COCO Caption 1k test images and 5k test images in Table~\ref{tab:supp-coco-full-1k} and Table~\ref{tab:supp-coco-full-5k}, respectively.
We also report additional experiments on PVSE such as larger $K$ ($K=4$), a different negative mining strategy (semi-hard negative mining~\cite{schroff2015facenet}.
In the tables, although \pcme shows slightly worse R@1 results than PVSE K=2, \pcme outperforms PVSE K=2 in PMRP scores.

\begin{figure}
    \centering
    \includegraphics[width=\linewidth]{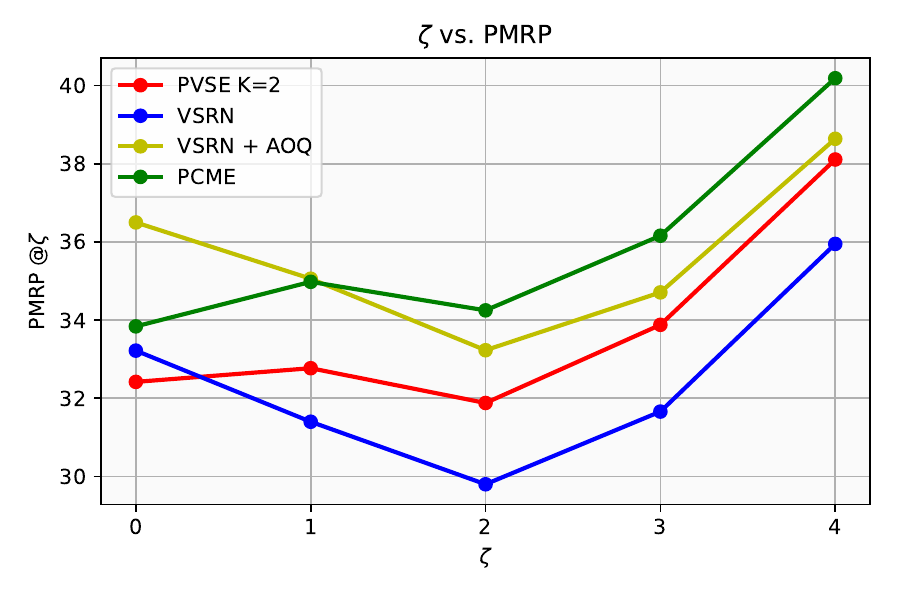}
    \caption{\small\textbf{PMRP by varying $\zeta$.} Plausible Match R-Precision scores for four methods with $\zeta = \{0, 1, 2\}$.}
    \label{fig:supp_pmrp_zeta}
\end{figure}

Also, we report PMRP scores of four methods (PVSE~\cite{song2019pvse}, VSRN~\cite{Li2019VSRN}, VSRN + AOQ~\cite{chen2020adaptive} and \pcme) by varying $\zeta$ for PMRP in Figure~\ref{fig:supp_pmrp_zeta}. In the figure, PMRP scores for VSRN and VSRN + AOQ are getting worse by increasing $\zeta$, in other words, theses method shows less coherence if we allow one missing or altering object class in the retrieved items. On the other hand, \pcme shows even increased performance with $\zeta > 0$, in other words, \pcme retrieves more plausible items than other methods.

\section{More uncertainty analysis}
\label{sec:supp-moreuncertainty}

Uncertainty estimation by \pcme brings interesting insights for the cross-modal retrieval tasks. In this section, we show additional uncertainty analysis for \pcme.

\subsection{Corruption vs. uncertainty in MS-COCO}
\label{sec:supp-coco-uncertainty-performance}

\begin{figure}
    \centering
    \includegraphics[width=.5\columnwidth]{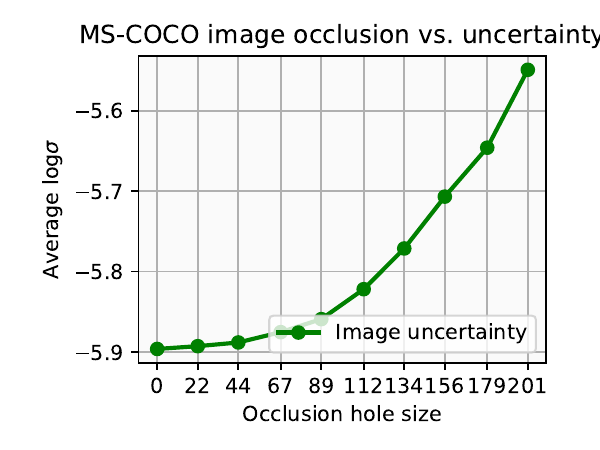}%
    \includegraphics[width=.5\columnwidth]{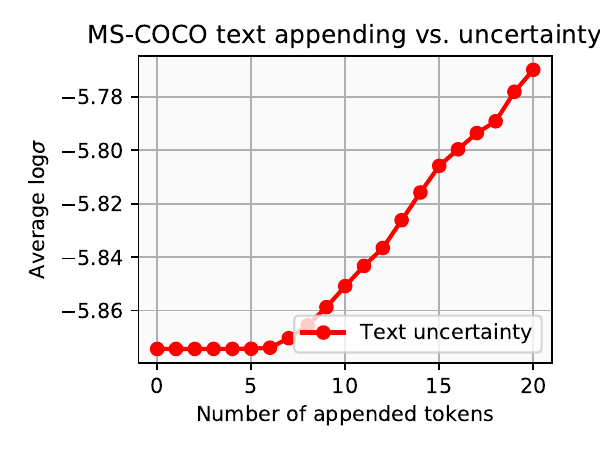}%
    \caption{\small\textbf{$\sigma$ captures ambiguity in COCO Caption.} Average $\log \sigma$ values at different ratios of erased pixels (for images) and appended $<$unk$>$ tokens (for captions).}
    \label{fig:coco-uncertainty-vs-corruption}
\end{figure}

As Figure~\textcolor{red2}{7}, 
we illustrate the uncertainty level by varying corruption levels on pixels and words in Figure~\ref{fig:coco-uncertainty-vs-corruption}. The left figure shows the uncertainty levels against occluded pixels. As we expected, more occlusion leads to higher uncertainty. The right figure shows the uncertainty levels against the number of appended $<$unk$>$ tokens. 

\subsection{Frequent words for each uncertainty bin}
\label{sec:supp-frequentword-per-uncertainty}

\begin{figure*}
    \centering
    \includegraphics[width=.2\linewidth]{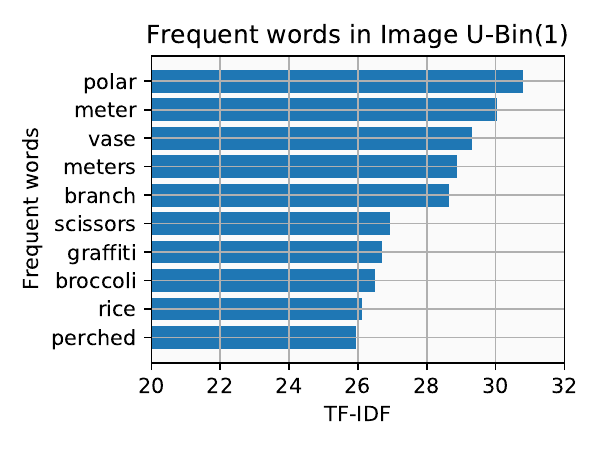}%
    \includegraphics[width=.2\linewidth]{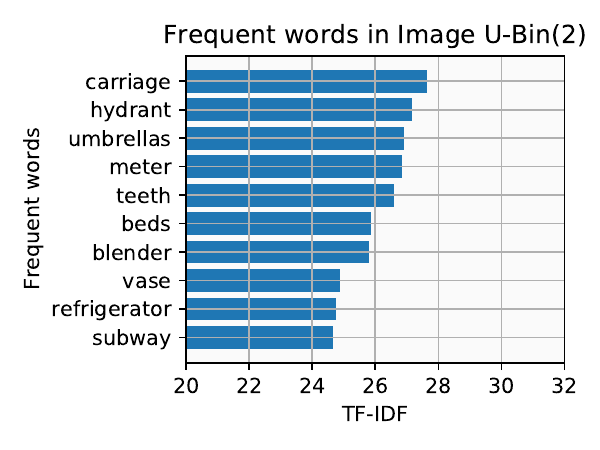}%
    \includegraphics[width=.2\linewidth]{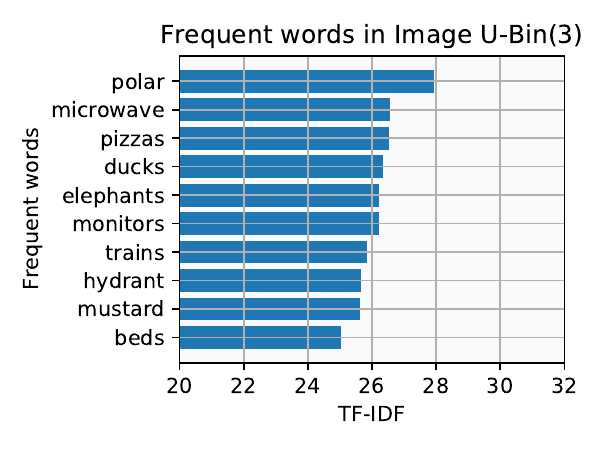}%
    \includegraphics[width=.2\linewidth]{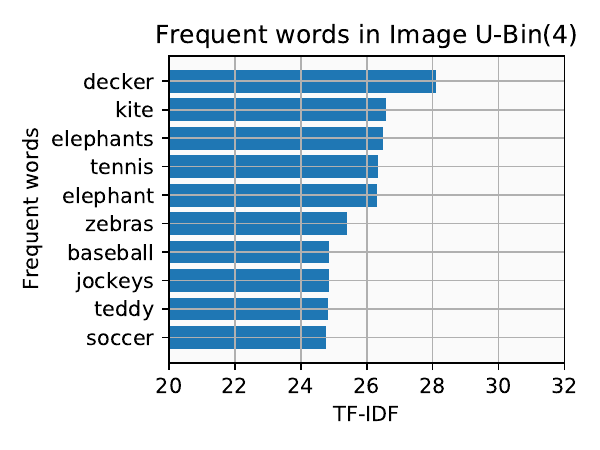}%
    \includegraphics[width=.2\linewidth]{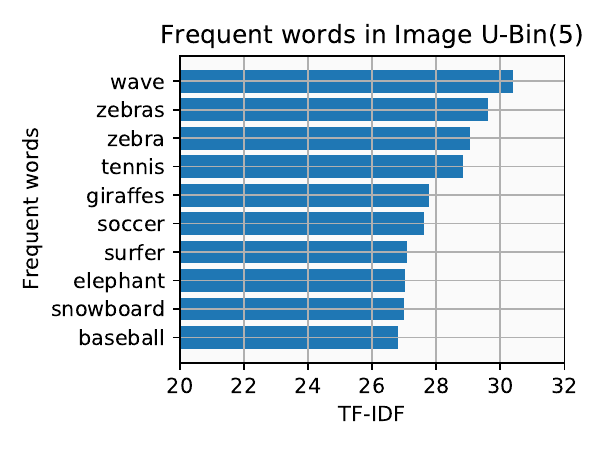}%
    
    \includegraphics[width=.2\linewidth]{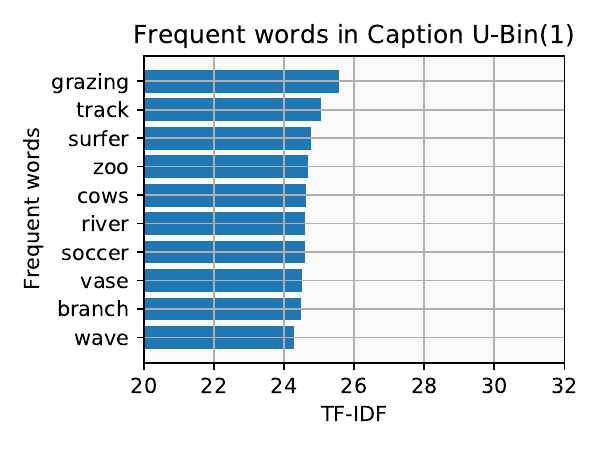}%
    \includegraphics[width=.2\linewidth]{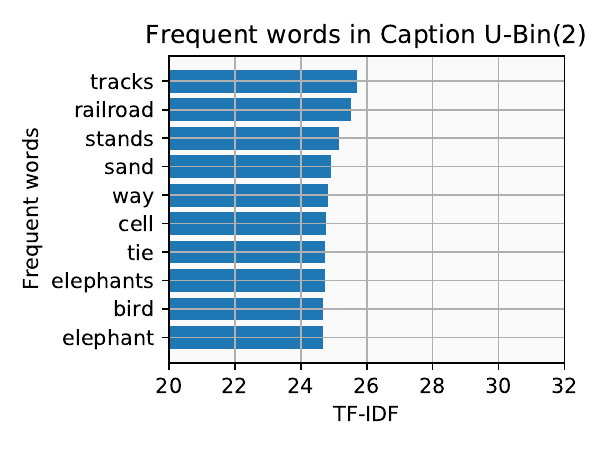}%
    \includegraphics[width=.2\linewidth]{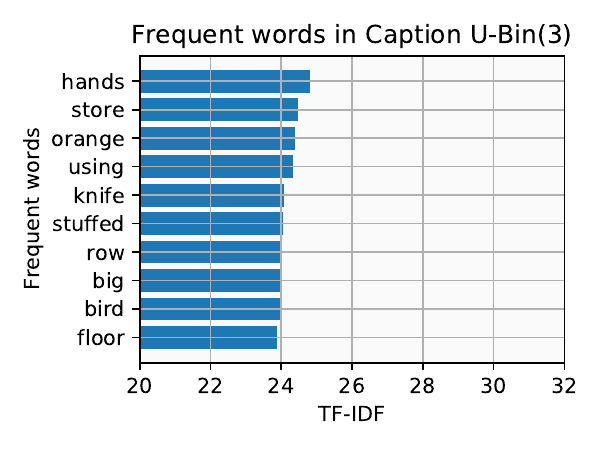}%
    \includegraphics[width=.2\linewidth]{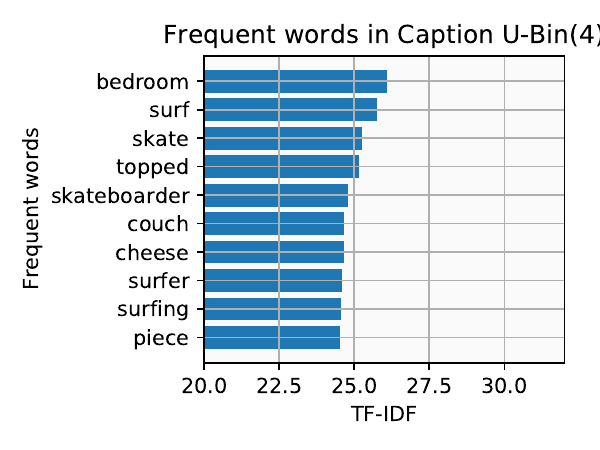}%
    \includegraphics[width=.2\linewidth]{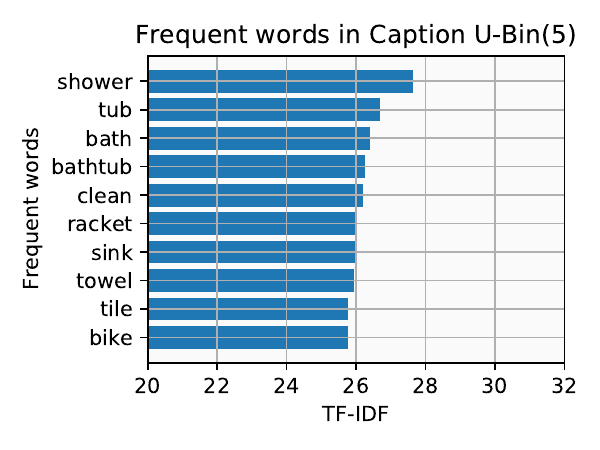}%
    \caption{\small\textbf{Frequent words in each uncertainty bin.} Term frequency–inverse document frequency (TF-IDF) sorted word frequencies are shown for each uncertainty bin (U-Bin, ascending order) for image (upper row) and caption (bottom row) modalities.}
    \label{fig:supp-uncertainty-tfidf}
\end{figure*}

Figure~\ref{fig:supp-uncertainty-tfidf} shows the frequent words per each uncertainty bin. We use term frequency–inverse document frequency (TF-IDF) as the frequent counter, defined as follows:
\begin{equation}
    \text{TF-IDF}(i) = (1 + \log n_i) \log \frac{N}{n_i},
\end{equation}
where $N$ is the number of total captions, and $n_i$ is the number of captions which contain word $i$. For the image word frequency, we use their ground truth captions for computing TF-IDF scores.

\subsection{Example uncertain samples}
\label{sec:supp-example-uncertain-samples}

\begin{figure*}
    \centering
    \includegraphics[width=\linewidth]{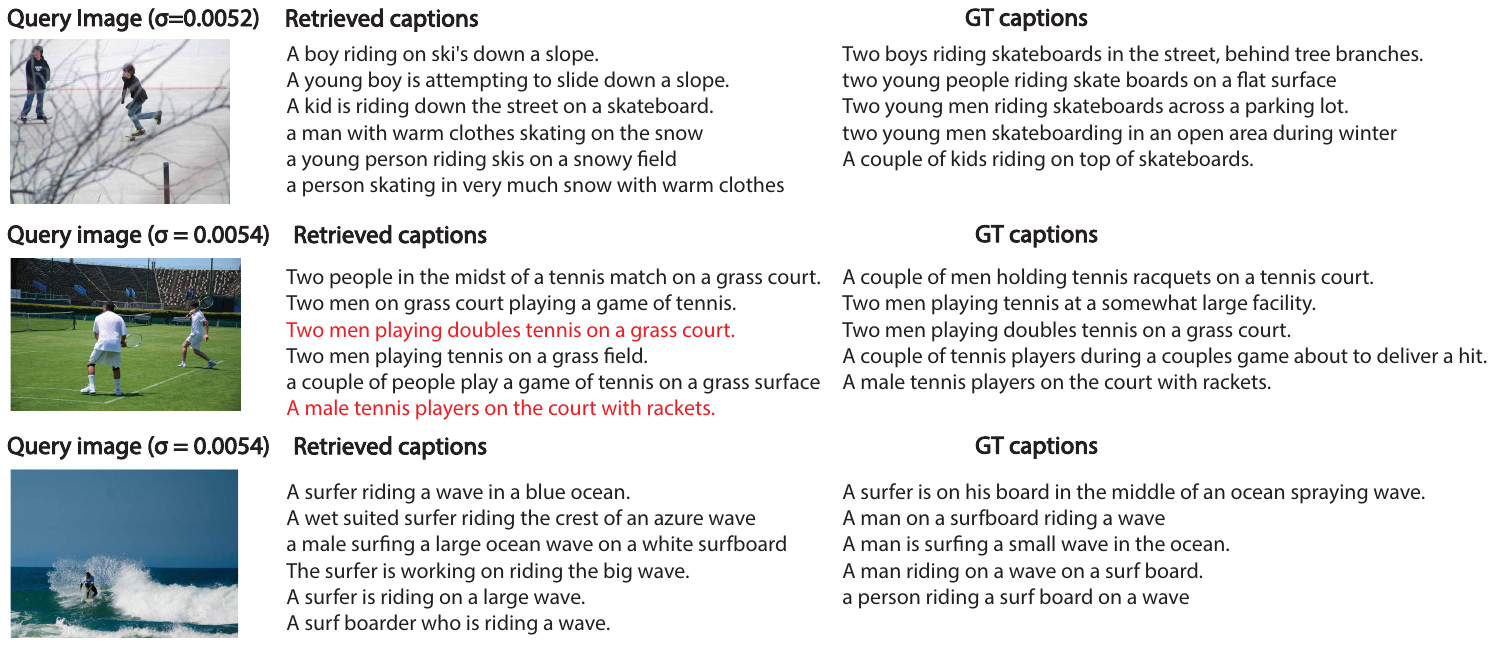}
    \caption{\small\textbf{Uncertain image examples.} Highly uncertain images, retrieved captions by \pcme, and their ground truth captions are shown.}
    \label{fig:supp-uncertain-images}
\end{figure*}

\begin{figure*}
    \centering
    \includegraphics[width=\linewidth]{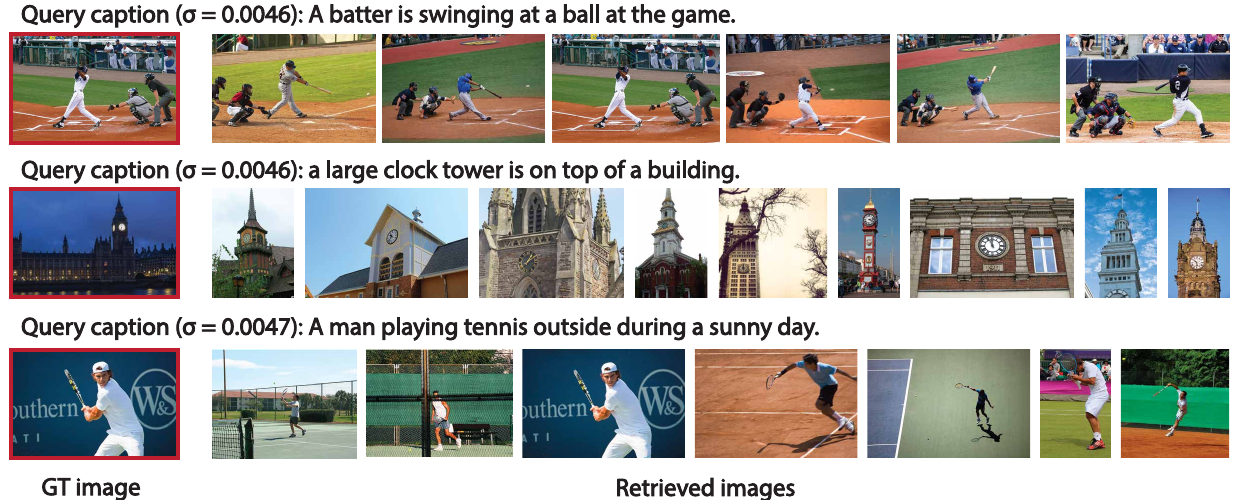}
    \caption{\small\textbf{Uncertain caption examples.} Highly uncertain captions, retrieved images by \pcme, and their ground truth image are shown.}
    \label{fig:supp-uncertain-captions}
\end{figure*}

We visualize the uncertain images and captions, and their corresponding retrieved items in Figure~\ref{fig:supp-uncertain-images} and Figure~\ref{fig:supp-uncertain-captions}. Interestingly, the retrieved captions and images are plausible results for the given query items. These qualitative results also show how the Recall@1 measure is noisy, and the proposed Plausible Match R-Precision (PMRP) is a more plausible and reliable measure to compare different retrieval methods.

\clearpage
\clearpage

\end{document}